\documentclass[journal]{IEEEtran}

\usepackage{cite}
\usepackage{amsmath,amssymb,amsfonts}
\usepackage{algorithmic}
\usepackage{graphicx}
\usepackage{textcomp}
\usepackage{xcolor}
\usepackage{cite}
\usepackage{booktabs}
\usepackage{subcaption}
\usepackage{placeins}
\usepackage{stfloats}
\usepackage{array}
\usepackage{url}

\begin{document}
\sloppy

\title{Overcoming `Physics Shock' in Earth Observation: A Heteroscedastic Uncertainty Framework for PINN-based Flood Inference}

\author{Tewodros Syum Gebre\textsuperscript{1}, 
        Jagrati Talreja\textsuperscript{1}, 
        Matilda Anokye\textsuperscript{1}, 
        and Leila Hashemi-Beni\textsuperscript{1,2},~\IEEEmembership{Member,~IEEE}%
\thanks{\textsuperscript{1}T. S. Gebre, J. Talreja, M. Anokye, and L. Hashemi-Beni are with the Built Environment Department, College of Science and Technology, North Carolina A\&T State University, Greensboro, NC 27405 USA (e-mail: tsgebre@ncat.edu; jtalreja@ncat.edu; manokye@ncat.edu; lhashemibeni@ncat.edu).}%
\thanks{\textsuperscript{2}L. Hashemi-Beni is also with the United Nations University Institute for Water, Environment and Health, Richmond Hill, ON, Canada.}%
\thanks{Corresponding author: Leila Hashemi-Beni.}}

\maketitle

\begin{abstract}
Rapid and accurate flood extent mapping from Remote Sensing data, such as Synthetic Aperture Radar (SAR), is critical for operational disaster response, but standard Deep Learning models often produce physically impossible predictions due to a lack of hydrological constraints. While Physics-Informed Neural Networks (PINNs) attempt to address this by embedding governing laws directly into the loss function, their application to real-world remote sensing data frequently fails. Enforcing rigid spatial derivatives (e.g., the 2D Shallow Water Equations) onto unconditioned latent spaces attempting to fit noisy SAR speckle causes catastrophic gradient divergence, a phenomenon we term Physics Shock. In this paper, we propose a novel Uncertainty-Aware PINN framework tailored specifically for applied Earth Observation that addresses this instability. By integrating a dynamic Warm-Start protocol and modeling heteroscedastic aleatoric uncertainty via a negative log-likelihood objective, the network learns to dynamically relax physical constraints in regions of high sensor noise while strictly enforcing them in high-confidence areas. Evaluated on the Sen1Floods11 dataset, our probabilistic Attention-Gated FNO-UNet successfully stabilizes multi-objective optimization, achieving a +25\% relative improvement in Intersection over Union (IoU) compared to deterministic baselines. Furthermore, through Deep Ensembles, we successfully disentangle intrinsic sensor noise from out-of-distribution terrain ignorance, providing operational agencies with highly calibrated, physically consistent confidence bounds for robust disaster mitigation and real-time decision-making.
This article is currently under review in IEEE Journal of Selected Topics in Applied Earth Observations and Remote Sensing
\end{abstract}


\section{Introduction}

\subsection{The Operational Imperative and the SAR Bottleneck}
Rapid and accurate mapping of flood extents is an operational imperative for disaster response, risk mitigation, and humanitarian resource allocation. In this context, Synthetic Aperture Radar (SAR) systems, such as the Sentinel-1 constellation, have have proven to be the gold standard for Earth Observation \cite{jamali2024residual}. Unlike optical sensors, SAR provides all-weather, day-and-night imaging capabilities, penetrating dense cloud cover to deliver critical snapshots of inundated regions during peak storm events \cite{fawakherji2025deepflood}. However, unlocking the full potential of SAR data is fundamentally constrained by its complex radiometric properties. The coherent nature of radar illumination inherently produces speckle noise, a granular interference that degrades image quality and obscures underlying spatial features. Furthermore, the side-looking geometry of SAR sensors introduces severe geometric and radiometric distortions, including terrain layover, foreshortening, and radar shadow, particularly in topographically complex or densely urbanized environments \cite{amitrano2024flood,salem2022inundated,hashemi2021flood}. These artifacts are not merely removable defects but represent intrinsic, irreducible data noise, formally known as \textit{aleatoric uncertainty}. Consequently, deterministic pixel-wise classification of water bodies becomes highly ambiguous. When models are forced to make hard predictions on this corrupted data, they inevitably overfit to the noise, drastically reducing the reliability and operational utility of the resulting flood maps.

\subsection{The Limits of Pure Deep Learning}
In recent years, data-driven Deep Learning (DL) architectures, particularly Convolutional Neural Networks (CNNs) and Vision Transformers, have achieved remarkable success in remote sensing computer vision tasks \cite{fawakherji2024multi}. By leveraging hierarchical feature extraction, these models excel at recognizing complex spatial patterns, vastly accelerating the speed of large-scale flood segmentation compared to traditional thresholding methods. However, pure DL models operate as unconstrained `black boxes' that treat flood mapping purely as an exercise in radiometric pattern recognition, entirely divorced from the underlying physics of fluid dynamics. This abstraction represents a critical flaw when applied to operational hydrology. Without an understanding of topography or conservation laws, standard neural networks frequently generate physically impossible predictions. Common failure modes include predicting deep standing water on steep terrain gradients, generating fragmented, hydraulically disconnected pools, or yielding massive false-positive classifications in urban zones due to radar double-bounce effects. Ultimately, relying solely on perception without physical grounding renders these models structurally inadequate for mission-critical flood inference.

\subsection{Physics-Informed Neural Networks and the `Physics Shock'}

\begin{figure*}[!t]
    \centering
    \includegraphics[width=0.95\linewidth]{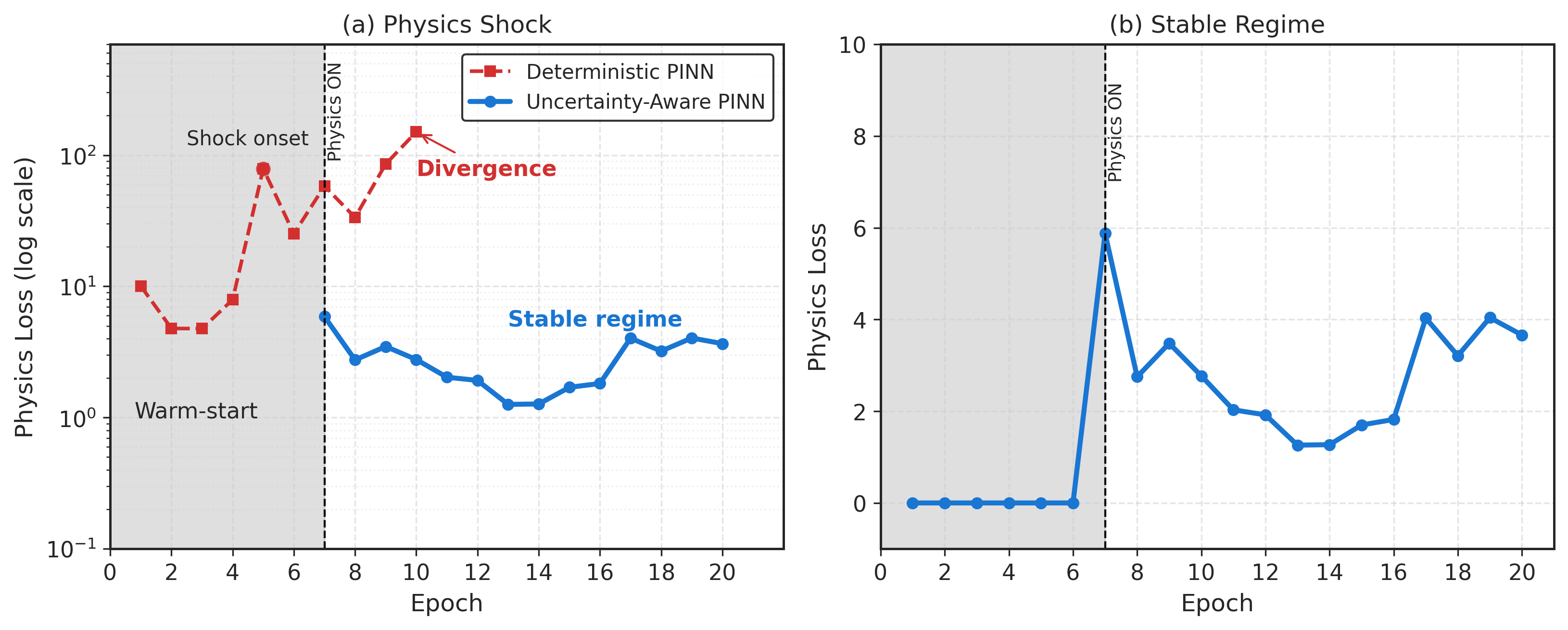}
    \caption{Conceptual illustration of the Physics Shock phenomenon. When rigid physical constraints are enforced on unconditioned latent representations of noisy SAR data, spatial derivatives amplify speckle, causing catastrophic divergence in physics loss (red). The proposed Uncertainty-Aware PINN mitigates this via a dynamic warm-start phase (gray), delaying physics enforcement until uncertainty is learned, resulting in stable optimization (blue).}
    \label{fig:physics_shock_phenomenon}
\end{figure*}

Physics-Informed Neural Networks (PINNs) offer a promising solution to this physical deficit by embedding governing hydrological laws, such as the 2D Shallow Water Equations (SWE), directly into the model's optimization objective \cite{tian2025physics,gebre2024ai}. While highly successful in simulated or pristine data environments, deploying PINNs on real-world Earth Observation data introduces a fatal optimization barrier \cite{gebre2025real,gebre2024integrated}. The physics loss heavily relies on computing spatial derivatives across the predicted outputs. We formally define this phenomenon as the \textbf{`Physics Shock'} (illustrated in Figure \ref{fig:physics_shock_phenomenon}). As demonstrated, when rigid physical constraints are immediately enforced upon an unconditioned latent space, the spatial derivatives massively amplify the intrinsic SAR speckle. This triggers a catastrophic gradient divergence, causing the physics loss to explode and halting convergence entirely. 

Current literature attempts to mitigate PINN gradient pathologies through purely mathematical dynamic weight balancing. However, these methods operate under the false assumption that the underlying training data is cleanly deterministic. By ignoring the intrinsic sensor physics, specifically the irreducible aleatoric noise of SAR, they force the model to apply strict physical rules to fundamentally untrustworthy pixels, invariably leading to model collapse. Our proposed framework addresses this issue by promoting stable training dynamics, as illustrated in Figure \ref{fig:physics_shock_phenomenon}.

\subsection{The Conceptual Leap: Uncertainty-Aware PINNs}
To bridge the profound gap between rigid mathematical physics and noisy Earth Observation sensors, we propose a paradigm shift: an Uncertainty-Aware PINN framework. Our approach uniquely combines Heteroscedastic Aleatoric Uncertainty modeling with a Dynamic Warm-Start protocol. Rather than treating all SAR pixels equally, the network is designed to simultaneously predict flood depth and its associated spatial variance.

In this paper, we present a novel, numerically stable framework for Physics-Guided flood inference, explicitly designed to transition PINNs from theoretical simulated environments to applied, operational deployment. Our core contributions are threefold:

\begin{itemize}
    \item \textbf{Mitigation of Physics Shock:} We introduce a Dynamic Warm-Start and stabilization protocol that prevents gradient divergence, which allows for the stable training of PINNs on highly noisy Earth Observation datasets.
    \item \textbf{Heteroscedastic Physics Integration:} We mathematically formulate a dual-objective loss function that leverages predicted aleatoric variance to balance rigid physical equations against sensor unreliability. This applied architecture achieves a +25\% relative improvement in Intersection over Union (IoU) compared to deterministic baselines.
    \item \textbf{Actionable Uncertainty Disentanglement:} Utilizing Deep Ensembles, we successfully disentangle intrinsic data noise (aleatoric) from out-of-distribution model ignorance (epistemic uncertainty). This provides operational disaster response agencies with highly calibrated, physically interpretable reliability metrics, bridging the gap between deep learning outputs and real-world risk assessment.
\end{itemize}

To support open science, the source code and model weights are made publicly available at \url{https://github.com/tsgebre/Flood_Physics_Guided_DL.git}.

\section{Related Work}

\subsection{Deep Learning for SAR Flood Mapping}

Historically, flood extent mapping from SAR imagery relied heavily on amplitude thresholding and change detection techniques \cite{Giustarini_2013,Martinis_2009,fawakherji2023multichannel,blay2024flood}. While computationally efficient, these classical methods are highly susceptible to radiometric artifacts, wind-induced surface roughening, and variations in incidence angles. Recently, the paradigm has decisively shifted toward Deep Learning (DL), with Convolutional Neural Networks (CNNs) like the U-Net and modern Vision Transformers (e.g., SegFormer) dominating the field \cite{Ronneberger_2015,Ding_2022}. These architectures have demonstrated exceptional capability in hierarchical feature extraction, implicitly mitigating SAR speckle and improving spatial contiguity in flood masks. Large-scale benchmark datasets, such as Sen1Floods11, have further catalyzed the development of these rapid, data-driven segmentation models \cite{Bonafilia_2020}. However, a fundamental and unsolved gap remains: pure DL architectures operate as unconstrained `black boxes.' They learn statistical correlations between radiometric signatures and annotated masks without any grounding in hydrological reality. Consequently, they are notoriously prone to generating physically impossible predictions, such as mapping deep standing water on steep terrain gradients or producing isolated, hydrodynamically disconnected pools. Furthermore, they frequently suffer from severe false positives in complex urban environments due to radar double-bounce effects \cite{Mason_2010}, highlighting the limitations of relying solely on perceptual pattern recognition for Earth System modeling.

\subsection{Physics-Informed Neural Networks in Geoscience}

To address the physical deficit of pure data-driven models, Physics-Informed Neural Networks (PINNs) have emerged as a transformative approach in the computational geosciences \cite{Karniadakis_2021,gebre2024ai,gebre2025real,gebre2025smart,gebre2024integrated}. By embedding governing differential equations, such as the Navier, Stokes equations or the Shallow Water Equations (SWE), directly into the loss function, PINNs constrain the neural network to produce solutions that respect fundamental conservation laws \cite{Raissi_2019,Mao_2020}. Despite their immense success in pristine simulated environments and controlled physical modeling, scaling PINNs to noisy, real-world Earth Observation data has proven notoriously difficult. The primary barrier is the multi-objective optimization conflict between the data-driven loss and the physics-guided residuals. Because physical constraints rely heavily on spatial derivatives, they act as high-pass filters. When naively applied to noisy sensor data, these derivatives amplify high-frequency artifacts (e.g., SAR speckle), leading to catastrophic gradient divergence, a phenomenon we term the `Physics Shock.'

To mitigate these gradient pathologies, researchers have developed sophisticated mathematical techniques, such as dynamic weight balancing and Neural Tangent Kernel (NTK)-guided weighting \cite{Wang_2021,Wang_2022}. While these methods effectively balance competing gradient magnitudes, they harbor a critical flaw: they implicitly treat all input data as deterministic ground truth. They mathematically balance the loss terms but entirely ignore the sensor physics. By forcing rigid physical equations onto fundamentally corrupted or ambiguous pixels without accounting for intrinsic sensor unreliability, these math-centric stabilization techniques inevitably fail when deployed on raw, uncalibrated satellite imagery.

\subsection{Uncertainty Quantification in Remote Sensing}

Recognizing the inherent noise and ambiguity in satellite observations, the remote sensing community has increasingly adopted Uncertainty Quantification (UQ) frameworks to improve model reliability. Techniques spanning Bayesian Deep Learning, Monte Carlo Dropout, and more recently, Evidential Deep Learning, have been successfully adapted to generate pixel-wise confidence intervals for land cover and flood segmentation \cite{Pandey_2024,Shiraishi_2025}. These models provide critical operational value by explicitly indicating where the network is uncertain, effectively flagging ambiguous regions such as dense vegetation, cloud cover, or radar shadow.

However, current UQ applications in remote sensing remain purely data-driven and passive. While these models successfully quantify uncertainty, they do not utilize physical laws to actively reduce or negotiate that uncertainty. The gap lies in the lack of a functional feedback mechanism between the sensor's confidence and the physical constraints of the environment. Our work directly addresses this gap. The primary differentiation axis of our proposed framework is the creation of an active feedback loop: we employ heteroscedastic aleatoric uncertainty not merely as a passive output metric, but as a dynamic gating mechanism within the optimization loop. By scaling the physics loss inversely with the predicted data variance, our model intelligently relaxes rigid physical rules in noisy regions while strictly enforcing them in clean areas, seamlessly bridging the gap between probabilistic perception and hydrodynamic reality.

\section{Methodology}

To systematically address the limitations of standard physics-informed architectures in noisy remote sensing environments, we propose a progressive algorithmic framework. While our experimental ablation study (Section IV) evaluates several intermediate deterministic models, this section details the complete mathematical and structural formulation leading to our final proposed State-of-the-Art (SOTA) model: the Uncertainty-Aware PINN.

\subsection{Problem Formulation and Baseline Physics Integration}

The foundational objective of our framework is to map heterogeneous remote sensing inputs, specifically SAR backscatter and high-resolution Digital Elevation Models (DEM), to a continuous spatial field representing flood inundation depth. In a standard deterministic Deep Learning formulation, the network predicts a scalar depth value $\mu(x)$ for every pixel $x$, optimizing a purely data-driven loss (e.g., Mean Squared Error, $\mathcal{L}_{MSE}$) against ground truth labels.

To move beyond this unconstrained perceptual mapping, Physics-Informed Neural Networks (PINNs) embed soft constraints derived from the steady-state 2D Shallow Water Equations (SWE) directly into the optimization landscape. Rather than demanding a perfect numerical simulation, the physics loss $\mathcal{L}_{phys}$ acts as a regularizer, penalizing predictions that are physically impossible. This loss is composed of two primary physical residuals.

First, we enforce \textbf{Mass Conservation (Continuity)}. Assuming a steady state, the continuity equation requires that $\nabla \cdot (\mu(x) \mathbf{v}) = 0$, where $\mathbf{v} = (u, v)$ represents the local velocity field vector, approximated via a proxy Manning's equation based on local depth and slope. The mass conservation loss penalizes the divergence of this flux, ensuring water is neither arbitrarily created nor destroyed:

\begin{equation}
\mathcal{L}_{mass} = \frac{1}{N_{water}} \sum_{i=1}^{N} \left[ \left( \frac{\partial (\mu u)}{\partial x} + \frac{\partial (\mu v)}{\partial y} \right)_i \cdot M_i \right]^2
\end{equation}

where $M_i \in \{0, 1\}$ is a binary mask ensuring constraints are only enforced within the inundated domain, and $N_{water}$ is the total number of water pixels.

Second, we enforce \textbf{Water Surface Elevation (WSE) Smoothness}. Under the hydrostatic assumption characteristic of standing floodwaters, the spatial gradient of the absolute water surface elevation should approach zero. We define WSE as the sum of the predicted water depth $\mu(x)$ and the underlying bare-earth elevation $Z(x)$ from the DEM:

\begin{equation}
\mathcal{L}_{smooth} = \frac{1}{N_{water}} \sum_{i=1}^{N} \left[ \left( \frac{\partial (\mu + Z)}{\partial x} \right)_i^2 + \left( \frac{\partial (\mu + Z)}{\partial y} \right)_i^2 \right] \cdot M_i
\end{equation}

The total baseline physics loss is the unweighted sum of these residuals: $\mathcal{L}_{phys} = \mathcal{L}_{mass} + \mathcal{L}_{smooth}$.

\subsection{Overcoming ``Physics Shock'': The Dynamic Warm-Start Protocol}

The standard objective for PINNs is to minimize a joint loss function: $\mathcal{L}_{total} = \mathcal{L}_{data} + \lambda \mathcal{L}_{phys}$. However, applying this joint objective naively to real-world SAR data reliably results in catastrophic optimization failure. Because $\mathcal{L}_{phys}$ heavily relies on the computation of first- and second-order spatial derivatives via finite-difference approximations, it acts as a high-pass filter. When applied to an unconditioned, noisy latent space from epoch zero, these derivative operators massively amplify local SAR speckle, generating explosive gradients that entirely overwhelm the data-driven loss. We term this rapid gradient divergence the \textbf{`Physics Shock.'}

To circumvent this temporal instability, we introduce a \textbf{Dynamic Warm-Start Protocol}, governed by an epoch-dependent, piecewise weighting function $\lambda_{phys}(e)$:

\begin{equation}
\lambda_{phys}(e) =
\begin{cases}
0 & \text{if } e < E_{warm} \\
\lambda_{max} \cdot \min\left(1, \frac{e - E_{warm}}{E_{ramp}}\right) & \text{if } e \ge E_{warm}
\end{cases}
\end{equation}

This protocol splits training into two distinct regimes. During \textbf{Phase 1: Pure Perception ($e < E_{warm}$)}, the physics loss is entirely disabled ($\lambda = 0$). The network focuses exclusively on $\mathcal{L}_{data}$ to learn basic radiometric feature mappings and establish a sensible topographical representation. Once the latent space is stabilized, the network enters \textbf{Phase 2: Physics Injection ($e \ge E_{warm}$)}. The physics constraints are linearly ramped up over $E_{ramp}$ epochs, avoiding destructive gradient interference.

\subsection{Architectural Evolution: Attention-Gated FNO-UNet}

\begin{figure}
    \centering
    \includegraphics[width=.8\linewidth]{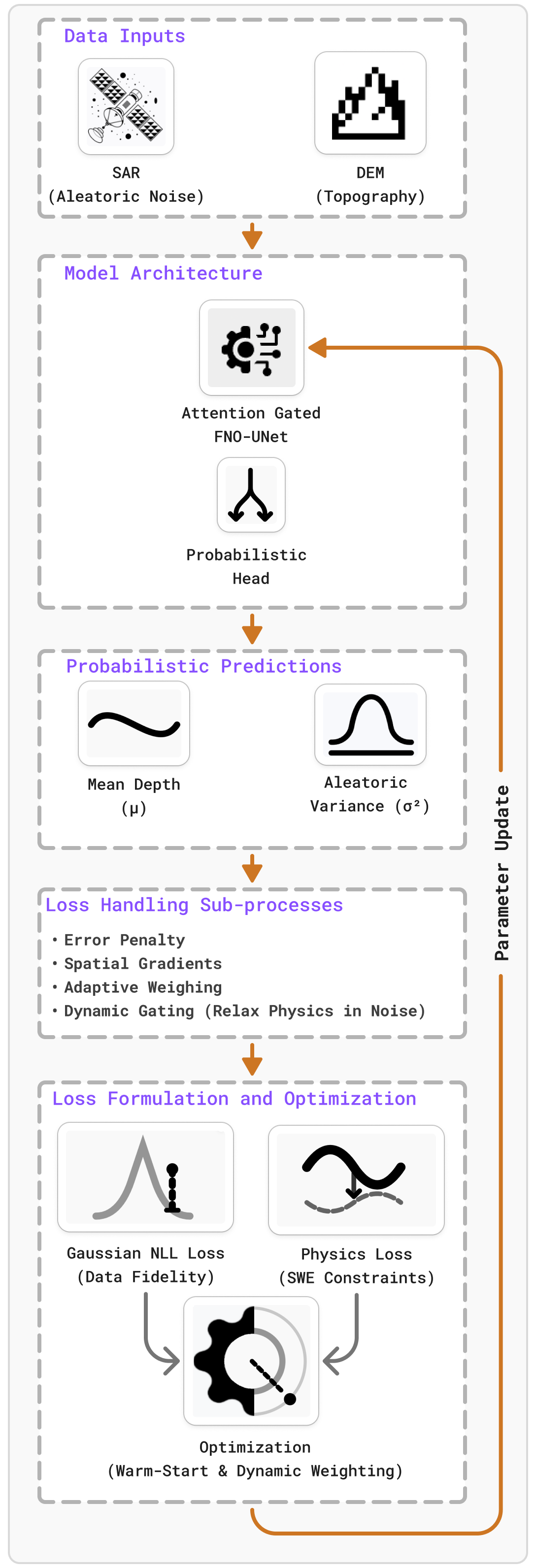}
    \caption{Schematic overview of the Uncertainty-Aware Physics-Informed Neural Network (PINN) architecture. The model integrates noisy SAR data and terrain DEM inputs through an Attention-Gated FNO-UNet backbone, followed by a probabilistic head that predicts both the mean depth ($\mu$) and aleatoric variance ($\sigma^2$). The predictions are guided by Gaussian NLL loss for data fidelity and physics-informed SWE constraints, with a warm-start optimization and dynamic weighting strategy to balance errors and relax physics in regions of high uncertainty.}
    \label{fig:arch}
\end{figure}

While temporal stabilization prevents immediate model collapse, accurate flood mapping requires harmonizing fine-grained local boundaries with macroscopic hydrological connectivity. Standard Convolutional Neural Networks (CNNs) rely on localized receptive fields, which systematically fail to capture the long-range spatial dependencies inherent to fluid dynamics. To bridge this gap without simply increasing network depth (which scales parameters inefficiently), we propose a hybrid architecture: the Attention-Gated FNO-UNet (illustrated in Figure~\ref{fig:arch}).

The core of this architecture is the Fourier Neural Operator (FNO) bottleneck. By parameterizing the integral kernel directly in Fourier space via Spectral Convolutions, the FNO bottleneck evaluates global spatial dependencies and continuous physical fields in a single forward pass. This guarantees that flood predictions in one region are hydrodynamically informed by distant topographical features.

Conversely, the decoding pathway must reconstruct precise, high-resolution inundation maps from SAR imagery notoriously corrupted by speckle noise. To prevent noise from bypassing the bottleneck, we integrate Attention Gates into the U-Net's skip connections. These gates dynamically compute spatial weighting masks, allowing the network to actively suppress highly noisy encoder features (e.g., severe speckle over open water) and amplify salient boundary signals (e.g., sharp topographic gradients).

\subsection{Modeling Heteroscedastic Aleatoric Uncertainty}

Even with a robust architecture and warm-start protocol, enforcing rigid physics on fundamentally corrupted SAR pixels (e.g., radar shadows, layover) forces the model into an ambiguous state. To resolve this spatial conflict, we reframe the deterministic learning objective into a probabilistic one by introducing a Probabilistic Head (detailed in Figure~\ref{fig:arch}).

In the terminal layer, rather than outputting a single scalar $\mu(x)$, the network bifurcates to output the parameters of a probability distribution: a predicted mean depth $\mu(x)$ and a predicted spatial variance $\sigma^2(x)$. This models \textit{heteroscedastic aleatoric uncertainty}, intrinsic data noise that varies spatially across the image.

We optimize this probabilistic framework by minimizing the Negative Log-Likelihood (NLL). The primary data-driven loss function, $\mathcal{L}_{data}$, is formulated as:

\begin{equation}
\mathcal{L}_{data} = \frac{1}{N} \sum_{i=1}^{N} \frac{1}{2} \left( \frac{(y_i - \mu(x_i))^2}{\sigma^2(x_i)} + \log \sigma^2(x_i) \right)
\end{equation}

This formulation acts as a self-calibrating, dynamic spatial gating mechanism. The denominator $\sigma^2(x_i)$ acts as an adaptive, pixel-wise learned weight. If a specific region is intrinsically noisy, the network predicts a large variance $\sigma^2$, effectively down-weighting the error penalty and communicating to the optimizer that the sensor data here is untrustworthy. The second term, $\log \sigma^2(x_i)$, acts as a strict mathematical regularizer, preventing the network from predicting infinite variance everywhere. Through this elegant balance, the model learns to map its own blindness, dynamically relaxing both data and physics constraints in noisy regions while strictly enforcing them in clean, high-confidence areas.

\subsection{Disentangling Uncertainty via Deep Ensembles}

While the probabilistic head successfully captures intrinsic sensor noise (aleatoric uncertainty), it cannot account for the network's own ignorance when presented with unfamiliar, out-of-distribution topography. To achieve a comprehensive uncertainty profile, we employ Deep Ensembles.

By training an ensemble of $M$ independent models with different weight initializations, we utilize the Law of Total Variance to decompose the total predictive uncertainty for a given pixel $x$ into two distinct components:

\begin{equation}
\sigma^2_{total}(x) = \underbrace{\frac{1}{M}\sum_{m=1}^M \sigma^2_m(x)}_{\text{Aleatoric}} + \underbrace{\frac{1}{M}\sum_{m=1}^M (\mu_m(x) - \mu_*(x))^2}_{\text{Epistemic}}
\end{equation}

Here, the \textbf{Aleatoric Uncertainty} (first term) is calculated as the mean of the predicted variances across the ensemble, identifying regions of irreducible sensor noise. The \textbf{Epistemic Uncertainty} (second term) is calculated as the variance of the predicted means (where $\mu_*$ represents the overall ensemble mean), isolating regions where the models disagree due to a lack of training context. This explicit disentanglement allows us to precisely trace the source of model failures for operational deployment.

\section{Results}

\subsection{Dataset, Implementation, and Ablation Setup}

\subsubsection*{Dataset Configuration}

To rigorously evaluate the proposed Uncertainty-Aware PINN framework, we utilized the Sen1Floods11 dataset, a widely used benchmark for SAR-based flood segmentation covering major flood events across the world \cite{Bonafilia_2020}.Sen1Floods11 is a large-scale benchmark specifically designed to advance SAR-based flood mapping and semantic segmentation. It was chosen as ideal condidate to this implementation as it captures exceptional geographic and topographical diversity by encompassing remote sensing data from 11 major flood events across six continents, including regions in Bolivia, Colombia, Ghana, and India.

To isolate the performance gains derived from our architectural improvements versus those gained from raw data scaling, our experimental protocol employed a two-tiered dataset configuration. First, for all baseline comparisons and architectural ablation studies, the models were trained and evaluated exclusively on the carefully hand-annotated (hand-labeled) split of the Sen1Floods11 dataset. This ensured a strictly controlled baseline environment for fair comparison. Second, to evaluate the peak operational robustness of the Uncertainty-Aware PINN, the optimal model was subsequently trained on the comprehensive dataset, incorporating both the hand-annotated imagery and the significantly larger corpus of weakly-labeled data.

\subsubsection*{Implementation Details}

The dataset was partitioned into an 80\% training and 20\% validation split, utilizing a strictly geographic separation to prevent spatial autocorrelation leakage. All models were implemented in PyTorch and trained on a couple of CUDA-enabled NIVIDIA A6000 GPUs. Training was conducted using the AdamW optimizer with an initial learning rate of 1e-3 (with learning rate scheduler enabled) and a batch size of 4 to 8. To ensure fair convergence comparisons, all models were trained for exactly 70 epochs using a Cosine Annealing Warm Restarts scheduler ($T_0=10, T_{mult}=2$) and an early stopping feature. The complete source code and model weights for the Uncertainty-Aware PINN are publicly available on GitHub\footnote{\url{https://github.com/tsgebre/Flood_Physics_Guided_DL.git}}.

\subsubsection*{Experimental Setup and Ablation Rationale}

To rigorously evaluate the impact of our proposed methodological advances, we designed a comprehensive architectural ablation study. The structural and algorithmic evolution of the evaluated models is summarized in Table \ref{tab:model_summary}.

\begin{table*}[htbp]
\centering
\caption{\textbf{Architectural and Algorithmic Evolution.} Progressive development of model architecture, loss formulation, and optimization interventions across experimental stages. All models were evaluated on a strictly isolated geographic holdout split.}
\label{tab:model_summary}

\resizebox{\textwidth}{!}{
\begin{tabular}{@{}l|
>{\raggedright\arraybackslash}p{4cm}
>{\raggedright\arraybackslash}p{4cm}
>{\raggedright\arraybackslash}p{5cm}
l@{}}
\toprule
\textbf{Model Version} & \textbf{Architecture Updates} & \textbf{Loss Formulation} & \textbf{Key Innovations / Stabilization} & \textbf{Est. Parameters} \\
\midrule
\textbf{Baseline Deterministic PINN} & Hybrid FNO-UNet (Width 32, Modes 16) & $\mathcal{L}_{MSE} + \lambda \mathcal{L}_{SWE}$ & None. Suffers from severe ``Physics Shock'' due to gradient divergence. & $\sim 4$ M \\
\addlinespace
\textbf{Stabilized Deterministic PINN} & Hybrid FNO-UNet (Width 32, Modes 16) & $\mathcal{L}_{MSE} + \lambda_{phys}(e) \mathcal{L}_{SWE}$ & Dynamic Warm-Start protocol (5 epochs) and input normalization. & $\sim 4$ M \\
\addlinespace
\textbf{Deep Deterministic PINN} & Deep Residual FNO-UNet (Width 64, Modes 16) & $\mathcal{L}_{MSE} + \lambda_{phys}(e) \mathcal{L}_{SWE}$ & Increased depth and capacity. Failed to outperform smarter probabilistic models. & $\sim 45$ M \\
\addlinespace
\textbf{Uncertainty-Aware PINN} & Attention-Gated FNO-UNet (Width 64, Modes 16). Probabilistic Head ($\mu, \sigma^2$) & Gaussian NLL + SoftDice + $\lambda_{phys}(e) \mathcal{L}_{SWE}$ & Heteroscedastic Aleatoric Uncertainty gates physics loss. Ensembling. & $\sim 36$ M \\
\bottomrule
\end{tabular}
}

\vspace{2pt}
\begin{minipage}{0.95\textwidth}
\small \textit{Note:} Deep Ensembles ($M=5$) were applied post-hoc to the Uncertainty-Aware model solely for estimating epistemic uncertainty and were not used in baseline IoU comparisons.
\end{minipage}

\end{table*}

The study progresses through four distinct stages, each designed to isolate and test a specific scientific hypothesis:

\noindent
\textbf{Baseline Deterministic PINN}: This model serves as the foundational control. By naively integrating physical constraints without prior stabilization, it establishes the primary failure mode of standard PINNs in remote sensing, the `Physics Shock', caused by the catastrophic interference between noisy SAR gradients and rigid physical laws.

\noindent
\textbf{Stabilized Deterministic PINN}: To validate the efficacy of our proposed stabilization strategies, this iteration introduces the Dynamic Warm-Start protocol and rigorous data normalization. It tests the hypothesis that conditioning the latent space prior to physics injection is a strict prerequisite for numerical stability.

\noindent
\textbf{Deep Deterministic PINN:} A persistent assumption in deep learning is that performance bottlenecks, including those caused by noisy data, can be overcome by simply scaling up representational capacity. By upgrading the backbone to a deep Res-UNet-FNO, this model explicitly tests whether deeper convolutional architectures can organically resolve the aleatoric ambiguity inherent in SAR imagery.

\noindent
\textbf{Uncertainty-Aware PINN:} Finally, our proposed State-of-the-Art (SOTA) framework introduces the Probabilistic Head to predict heteroscedastic aleatoric variance. This stage is designed to conclusively prove that explicitly modeling sensor noise probabilistically, which allows the network to dynamically gate physics constraints, is the optimal and most stable solution for robust flood inference.

\subsection{Overcoming the `Physics Shock' in Multi-Objective Optimization}

\begin{figure*}[ht]
    \centering
    \includegraphics[width=.97\linewidth]{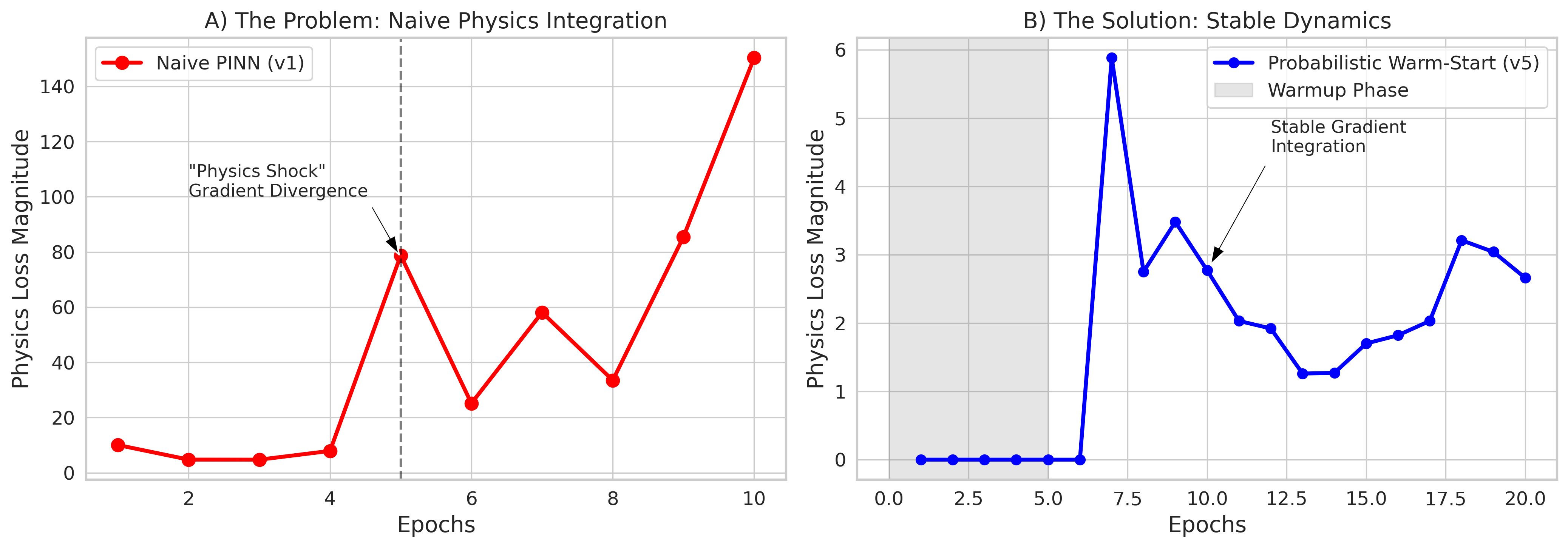}
    \caption{Training dynamics illustrating the `Physics Shock' and its mitigation. (A) In a baseline deterministic PINN, the immediate enforcement of Shallow Water Equation (SWE) constraints amplifies high-frequency SAR noise through spatial derivatives, causing the physics loss to rapidly diverge and destabilize training. (B) In contrast, the proposed uncertainty-aware framework with a dynamic warm-start delays physics injection, allowing the network to first learn a stable probabilistic representation of the data. This prevents gradient explosion and yields smooth, stable convergence of the physics loss.}
    \label{fig:shock}
\end{figure*}

\begin{figure*}[ht]
    \centering
    \includegraphics[width=.97\linewidth]{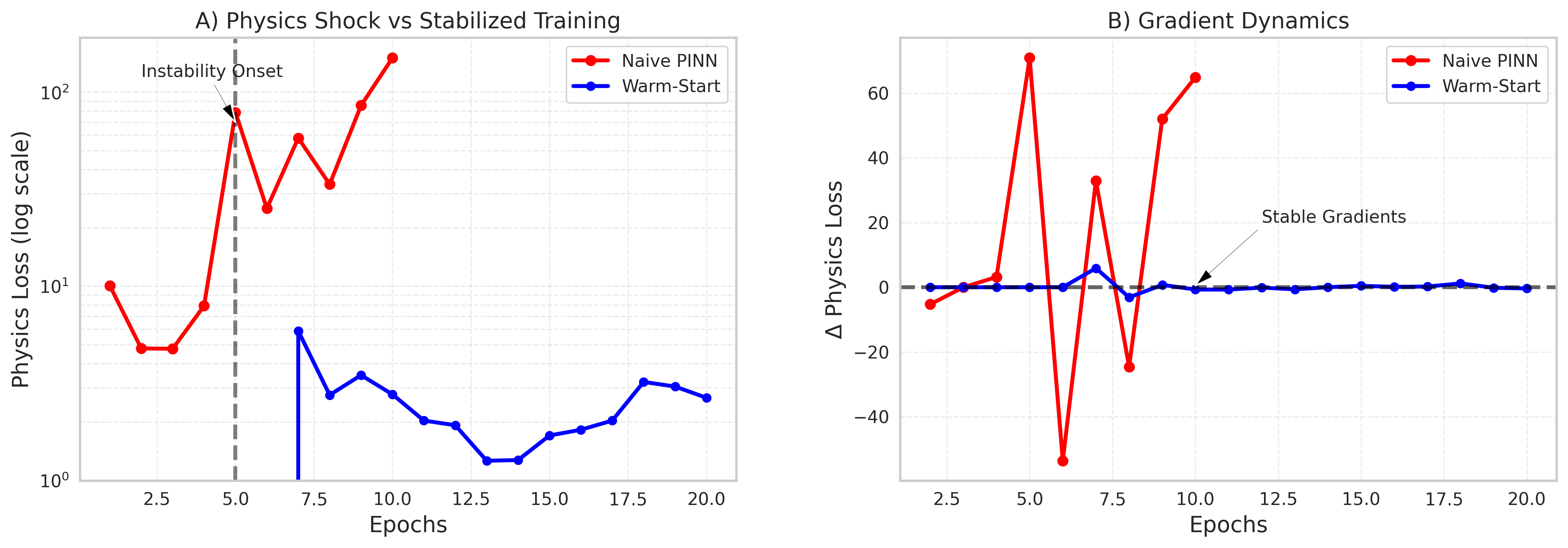}
    \caption{Training dynamics alternative perspectives with loss scaling and gradient behavior. (A) The physics loss trajectories from Fig.~\ref{fig:shock} are replotted on a logarithmic scale to emphasize relative changes across orders of magnitude. (B) The first-order difference of the physics loss ($\Delta$ loss) is shown to directly visualize gradient dynamics across epochs.}
    \label{fig:gradient}
\end{figure*}

The fundamental challenge of integrating the Shallow Water Equations (SWE) into a deep learning framework is vividly illustrated in the training dynamics (Figures \ref{fig:shock} and \ref{fig:gradient}). When physics constraints are naively applied from the first epoch alongside the data-driven loss, the network experiences a catastrophic optimization failure, which we term the `Physics Shock'. As seen in Figure \ref{fig:shock}A, the baseline deterministic PINN initially attempts to map the noisy SAR speckle. However, the spatial derivative operators within the SWE act as high-pass filters, amplifying this high-frequency noise and generating massive, conflicting gradients. The severity of this divergence becomes acutely apparent on a logarithmic scale (Figure \ref{fig:gradient}A), revealing abrupt, multiplicative increases in loss. Furthermore, plotting the first-order difference of the physics loss ($\Delta$ loss) directly visualizes this instability, exhibiting large, erratic fluctuations indicative of gradient amplification (Figure \ref{fig:gradient}B). Consequently, the physics loss rapidly explodes from an initial magnitude of $\sim$10 to over 150 within just five epochs, destroying the learned latent space and halting convergence.

In contrast, the stabilizing power of our proposed Uncertainty-Aware framework coupled with the Dynamic Warm-Start protocol is evident across all metrics. By entirely disabling the physics loss for the first $x$ epochs (5 epochs for the visualization case), the network is allowed to construct a stable, probabilistic mapping of the terrain and speckle noise. When the physics constraints are subsequently ramped up, the gradients do not explode. Instead, the physics loss smoothly stabilizes at a magnitude of $\sim$6 (Figure \ref{fig:shock}B). 

The alternative visualizations confirm this stabilization: the log-scale trajectory maintains consistently low-magnitude variations after activation (Figure \ref{fig:gradient}A), and the proposed warm-start strategy yields controlled, near-zero updates reflecting a stable gradient flow (Figure \ref{fig:gradient}B). These visual analyses strengthen the intervention that delaying physics injection until the network has mapped the aleatoric uncertainty is not merely an optimization trick, but a strict prerequisite for preventing gradient divergence and achieving stable multi-objective convergence in noisy Earth Observation domains.

\subsection{Quantitative Performance}

\begin{figure*}[!ht]
    \centering
    \includegraphics[width=.97\textwidth]{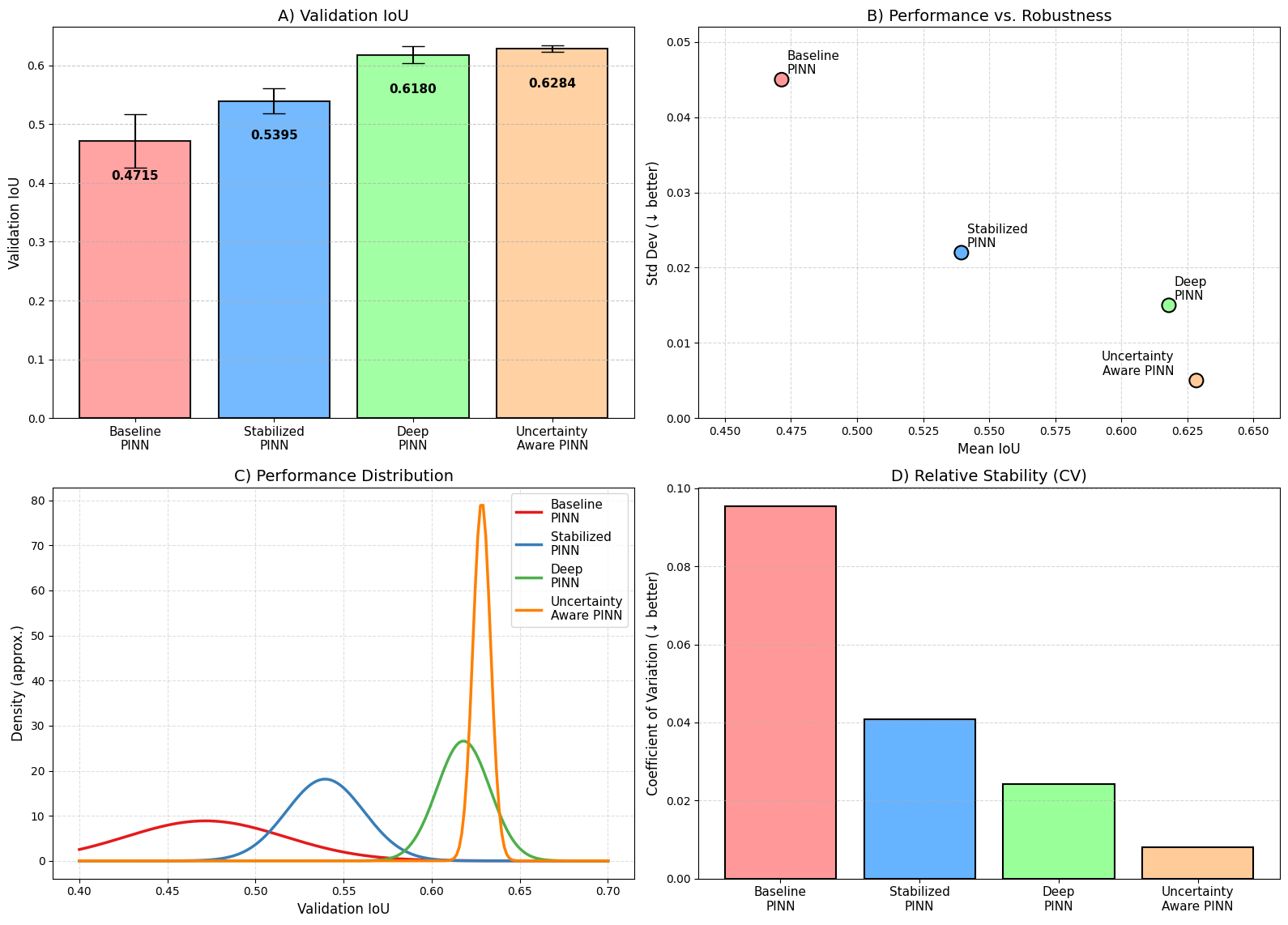}
    \caption{\textbf{Quantitative Performance Summary: Probabilistic vs. Deterministic Scaling.} \textbf{A)} Mean Validation IoU across ablated architectures, highlighting the +25\% relative improvement of the Uncertainty-Aware PINN over the baseline. Error bars denote $\pm$ 1 standard deviation. \textbf{B)} Performance vs. Robustness scatter plot demonstrating that the SOTA model is both the most accurate and the most stable (lowest standard deviation). \textbf{C)} Approximated Probability Density Function (PDF) of Validation IoU across ensemble runs, emphasizing the tight distribution of the probabilistic model. \textbf{D)} Relative Stability measured by the Coefficient of Variation, showing a substantial reduction in relative error variance for the proposed architecture.}
    \label{fig:quant_perf}
\end{figure*}

To isolate the drivers of our performance gains, we conducted a rigorous architectural ablation study, the results of which are visualized in Figure \ref{fig:quant_perf}. The unconstrained \textbf{Baseline Deterministic PINN}, representing a pure perceptual approach without physical priors or stabilization, achieved an Intersection over Union (IoU) of 0.4715 (Figure \ref{fig:quant_perf}A). This model struggled heavily with false positives in areas of dense speckle and radar shadow. Incorporating the warm-start protocol and data normalization yielded the \textbf{Stabilized Deterministic PINN}, which successfully prevented physics shock and improved the IoU to 0.5395 by enforcing basic hydrological consistency.

A common assumption in deep learning is that remaining performance bottlenecks can be overcome by simply increasing representational capacity. To test this hypothesis, we scaled the deterministic architecture to a \textbf{Deep Deterministic PINN} (utilizing a deep Res-UNet-FNO backbone). While this deeper model improved the IoU to 0.6180, it exhibited severe training instability and failed to fully resolve physical inconsistencies at complex flood boundaries. Conversely, our proposed \textbf{Uncertainty-Aware PINN} (SOTA), which maintains a shallower backbone but explicitly models heteroscedastic aleatoric uncertainty, achieved the highest overall performance with an IoU of 0.6284 (Figure \ref{fig:quant_perf}A). This represents a substantial absolute gain and a +25\% relative improvement over the baseline.

The critical relationship between accuracy and model reliability is highlighted in the performance versus robustness scatter plot (Figure \ref{fig:quant_perf}B) and the coefficient of variation chart (Figure \ref{fig:quant_perf}D). The Uncertainty-Aware PINN not only achieves the highest mean IoU but also exhibits the lowest standard deviation ($\pm$ 0.005), marking it as the most stable architecture. In contrast, while the deeper deterministic model improved average performance, its variance remained notably higher.

This reliability is further corroborated by the probability density function (PDF) plot (Figure \ref{fig:quant_perf}C), which illustrates the performance distribution across models. The SOTA model's distribution is significantly tighter and definitively shifted toward higher accuracy compared to the wider, lower distributions of the baseline models. Collectively, these quantitative metrics establish our core claim: `Smarter is better than Deeper.' In the context of SAR flood mapping, the primary bottleneck is not the network's representational capacity, but the inherently low signal-to-noise ratio of the data. Explicitly modeling this noise via a probabilistic head yields significantly higher returns and operational stability than blindly increasing convolutional depth.

\begin{figure*}[!t]
    \centering

    \begin{subfigure}{\linewidth}
        \includegraphics[width=\linewidth]{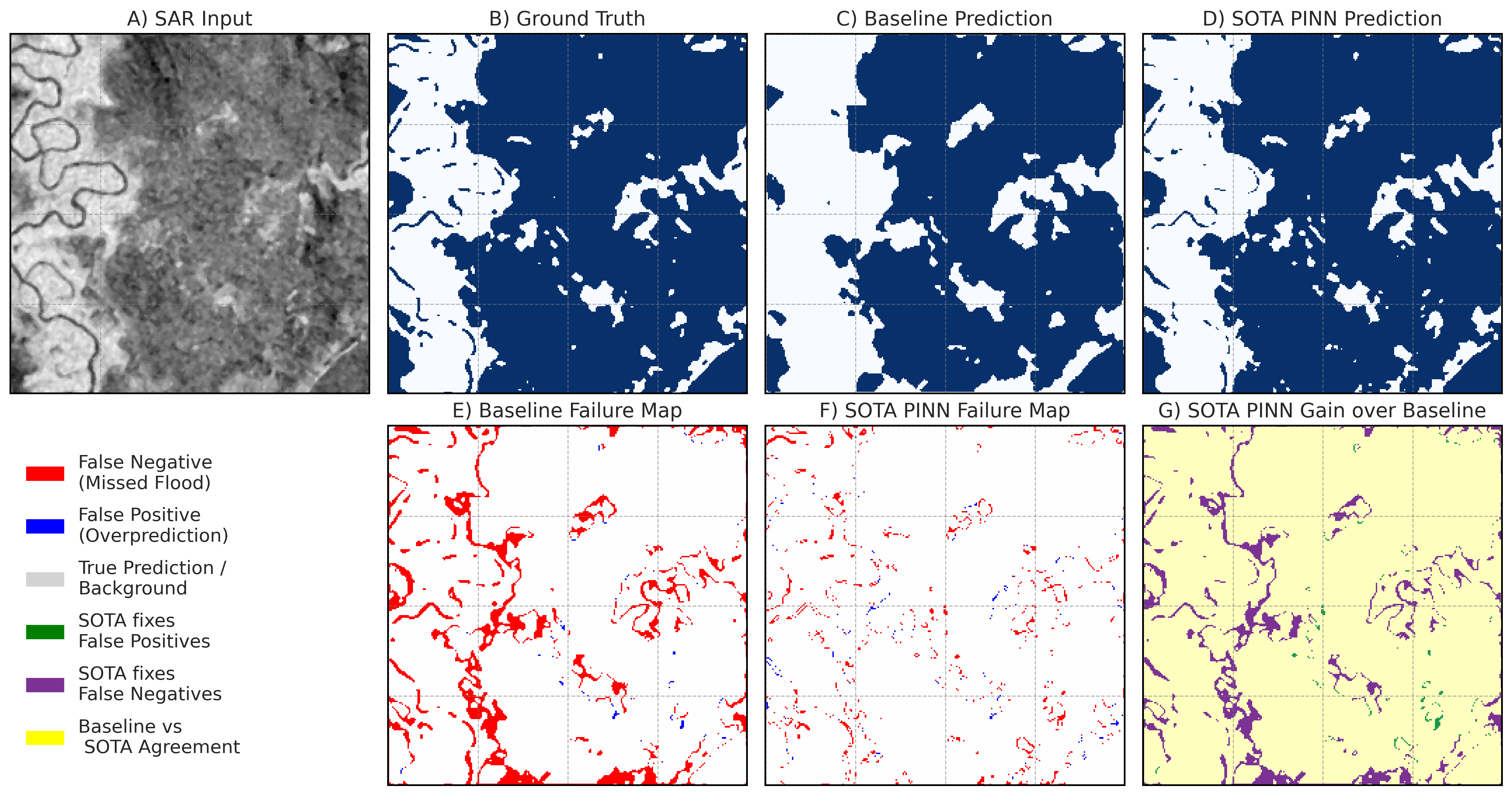}
        \caption{Bolivia (2018--02--15)}
    \end{subfigure}

    \vspace{0.2cm}

    \begin{subfigure}{\linewidth}
        \includegraphics[width=\linewidth]{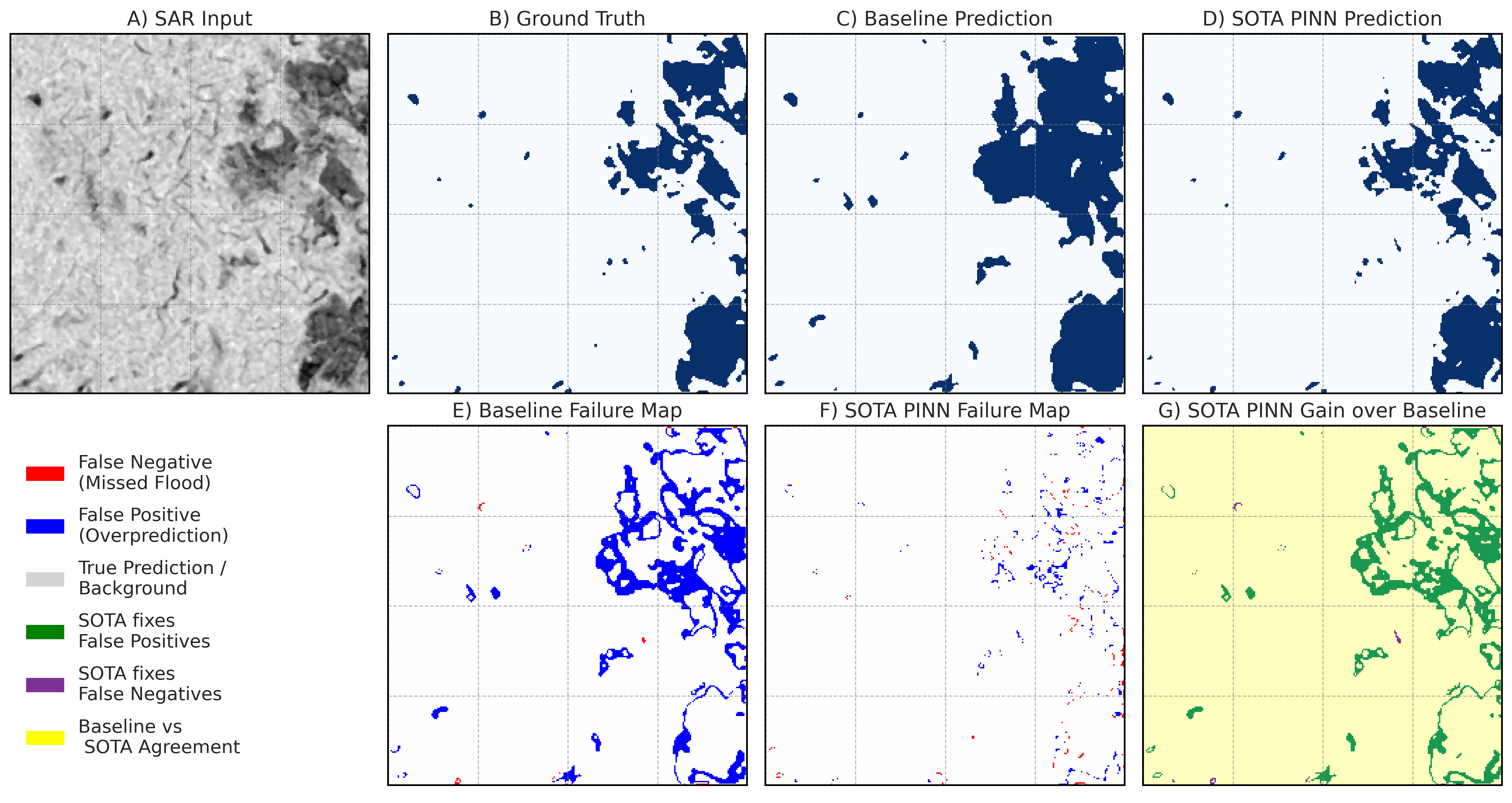}
        \caption{Colombia (2018--05--01)}
    \end{subfigure}

    \caption{\textbf{Qualitative flood mapping results (Part 1 of 2).}}
    \label{fig:inference1}
\end{figure*}

\begin{figure*}[!t]
    \centering

    \begin{subfigure}{\linewidth}
        \includegraphics[width=\linewidth]{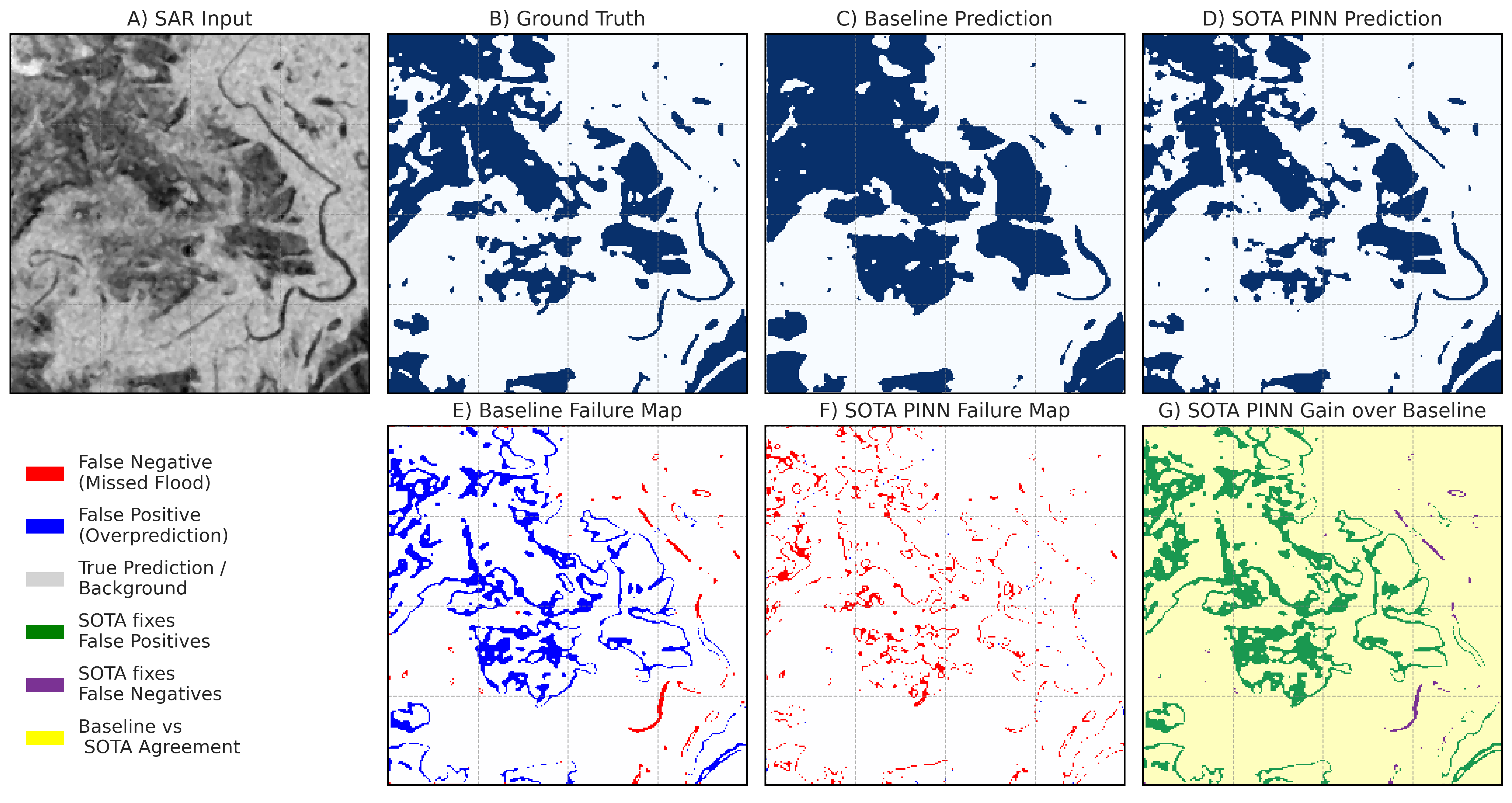}
        \caption{Ghana (2018--09--18)}
    \end{subfigure}

    \vspace{0.2cm}

    \begin{subfigure}{\linewidth}
        \includegraphics[width=\linewidth]{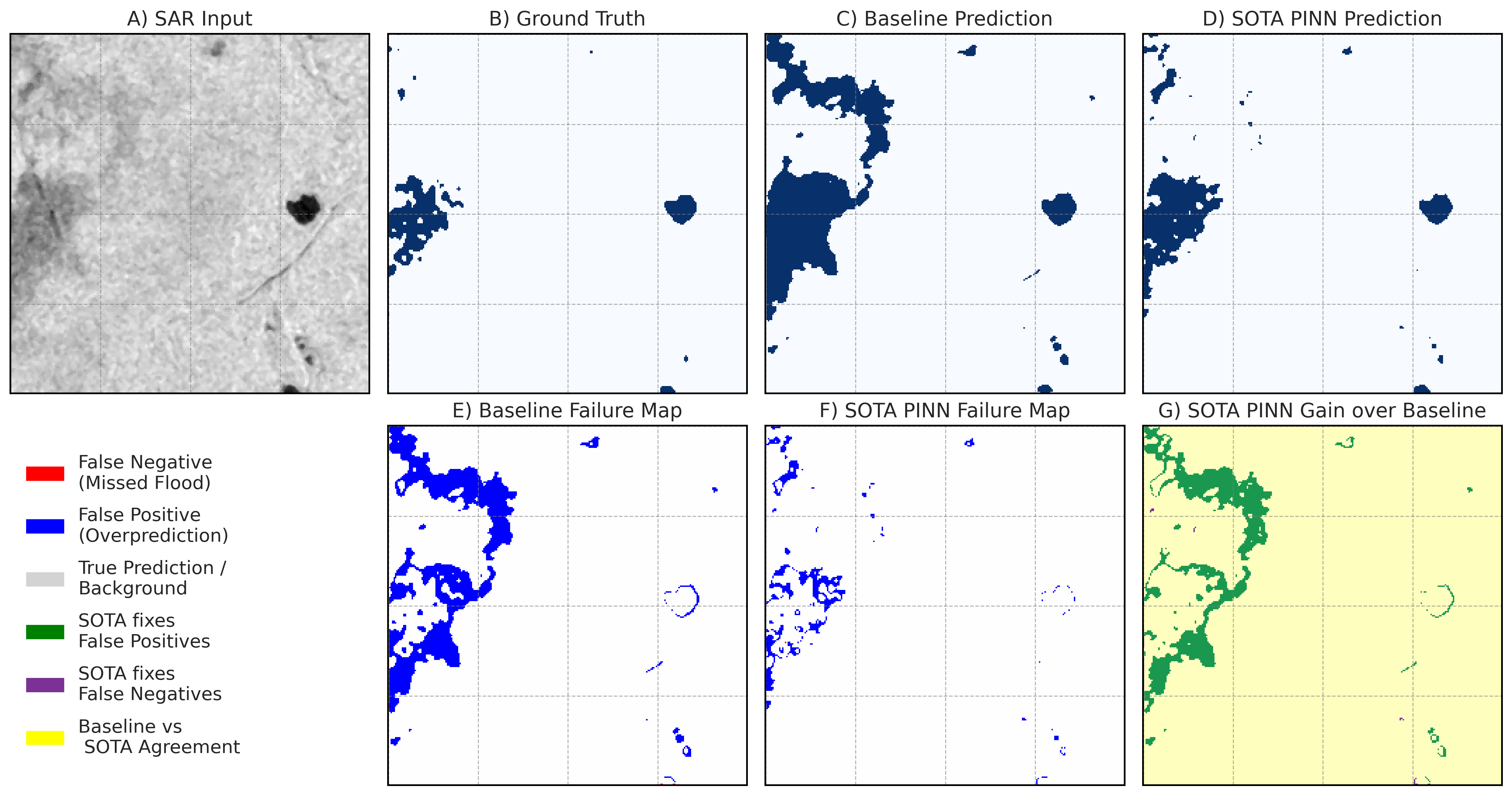}
        \caption{India (2016--08--12)}
    \end{subfigure}

    \caption{\textbf{Qualitative flood mapping results (Part 2 of 2).}}
    \label{fig:inference2}
\end{figure*}

The quantitative superiority is qualitatively supported by the visual predictions presented in Figures~\ref{fig:inference1}--\ref{fig:inference2}. The unconstrained Baseline Deterministic PINN (Figures~\ref{fig:inference1}--\ref{fig:inference2}C) struggles heavily with the unconditioned SAR data, hallucinating deep standing water in radar shadows and producing fragmented, hydrodynamically disconnected flood regions. These failure modes are further highlighted in the corresponding error map (Figures~\ref{fig:inference1}--\ref{fig:inference2}E), which shows pronounced false positives and false negatives across ambiguous backscatter zones. In contrast, the proposed Uncertainty-Aware PINN (Figures~\ref{fig:inference1}--\ref{fig:inference2}D) effectively suppresses these noise-induced artifacts, yielding more spatially coherent flood extents. The associated error visualization (Figures~\ref{fig:inference1}--\ref{fig:inference2}F) confirms a substantial reduction in both omission and commission errors. Furthermore, the improvement map (Figures~\ref{fig:inference1}--\ref{fig:inference2}G) clearly demonstrates regions of consistent error reduction, indicating where the SOTA model successfully corrects baseline failures. By dynamically relaxing constraints in ambiguous areas, the proposed model produces smoother and more physically consistent flood boundaries, which aligned with underlying topographic structure and visually supporting the reported +25\% IoU gain.

\subsection{Peak Operational Performance and Training Dynamics}

To fully evaluate the peak capabilities of the proposed uncertainty-aware framework, the optimal model configuration (Uncertainty-Aware PINN) was scaled to a broader data distribution, utilizing both the hand-labeled imagery and the extensive corpus of weakly-labeled samples from the Sen1Floods11 dataset (totaling 4,383 samples). 

\begin{figure*}[!htb]
    \centering
    \includegraphics[width=.85\textwidth]{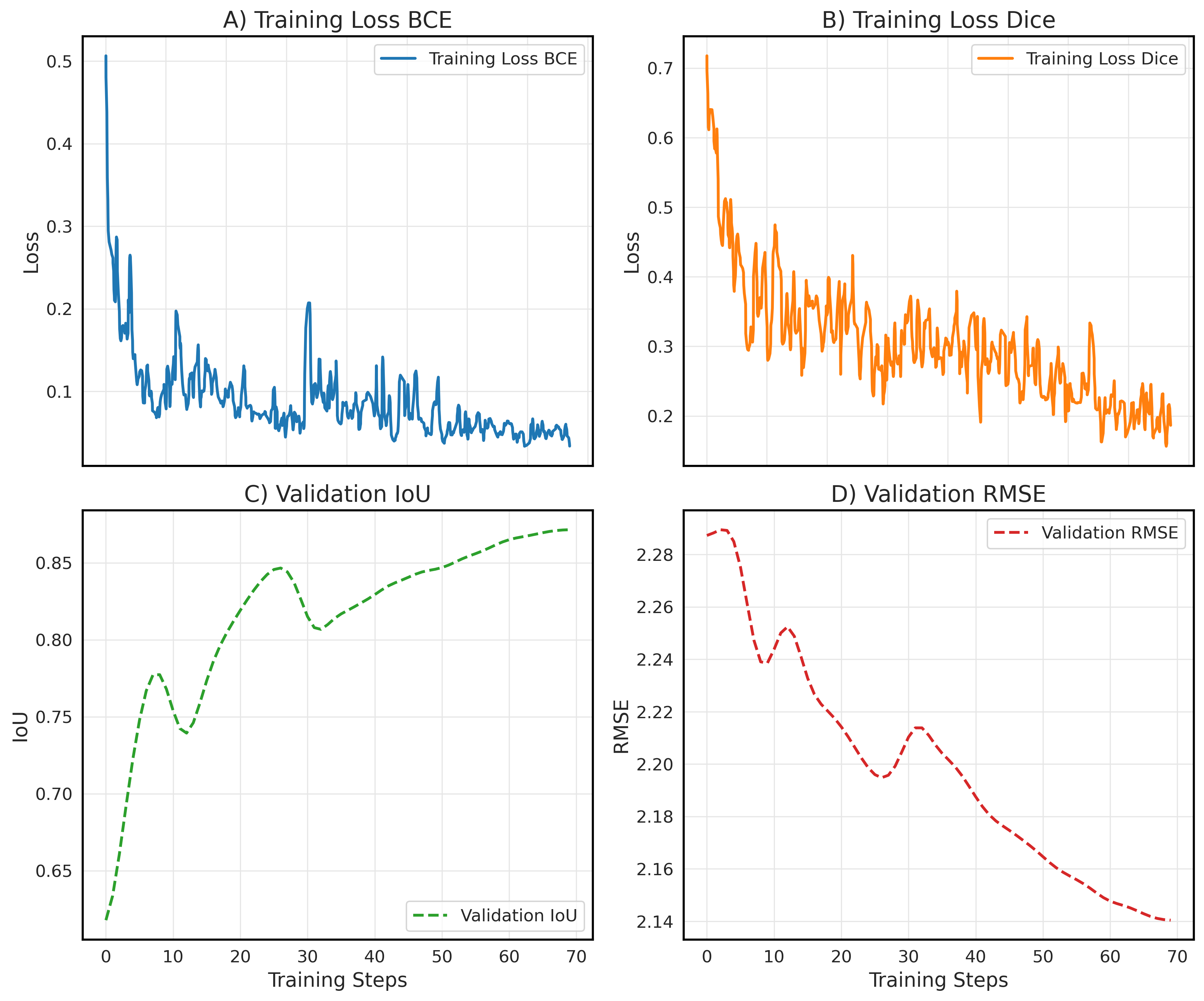}
    \caption{Training dynamics and validation performance during model training. \textbf{(A)} Training Loss (BCE), \textbf{(B)} Training Loss (Dice), \textbf{(C)} Validation Intersection over Union (IoU), and \textbf{(D)} Validation Root Mean Square Error (RMSE). Each plot shows the corresponding metric over training steps, with smoothed curves.}
    \label{fig:training_dynamics}
\end{figure*}

The stability and success of this scaling are explicitly captured in the training dynamics presented in Figure \ref{fig:training_dynamics}. As the network processes these extensive and diverse training signals, the loss components (Figure \ref{fig:training_dynamics}A-B) demonstrate smooth, stable convergence, entirely avoiding the gradient divergence typical of standard PINNs. 

The final aggregated quantitative performance across the validation holdout is detailed in Table \ref{tab:peak_performance}. The validation metrics highlight a significant leap in mapping accuracy: the model achieves an exceptional peak Intersection over Union (IoU) of 0.8562 alongside an F1-Score of 0.9225. Crucially, the model demonstrates a remarkable Precision of 0.9531, ensuring minimal false positives—a vital characteristic for operational disaster response where false alarms misallocate critical resources. Concurrently, the physical depth estimation error remains highly bounded, culminating in a Mean Absolute Error (MAE) of just 0.2595 meters and a Root Mean Square Error (RMSE) of 3.2823 meters. 

Yet, while these aggregated global metrics confirm the model's structural capacity to learn robust, physically compliant representations from noisy SAR data, they inherently obscure the local, pixel-wise negotiations occurring within the network. To understand the underlying mechanics enabling this success, we must examine the spatial distribution of the model's doubt.

\begin{table}[htbp]
    \centering
    \caption{\textbf{Peak Operational Performance Metrics.} Quantitative evaluation of the Uncertainty-Aware PINN scaled to the full multi-tier Sen1Floods11 dataset. Metrics evaluate both classification accuracy (segmentation) and physical consistency (depth estimation).}
    \label{tab:peak_performance}
    \begin{tabular}{l l}
        \toprule
        \textbf{Segmentation Metrics} & \textbf{Depth Estimation Metrics} \\
        \midrule
        Intersection over Union (IoU) & Root Mean Square Error (RMSE) \\
        0.8562                       & 3.2823 m \\
        F1-Score (Dice Coefficient)  & Mean Absolute Error (MAE) \\
        0.9225                       & 0.2595 m \\
        Precision                    &  \\
        0.9531                      &  \\
        Recall                       &  \\
        0.8939                      &  \\
        \bottomrule
    \end{tabular}
\end{table}

\subsection{Spatial Anatomy of Aleatoric Uncertainty}

\begin{figure*}[ht!]
    \centering
    \includegraphics[width=\textwidth]{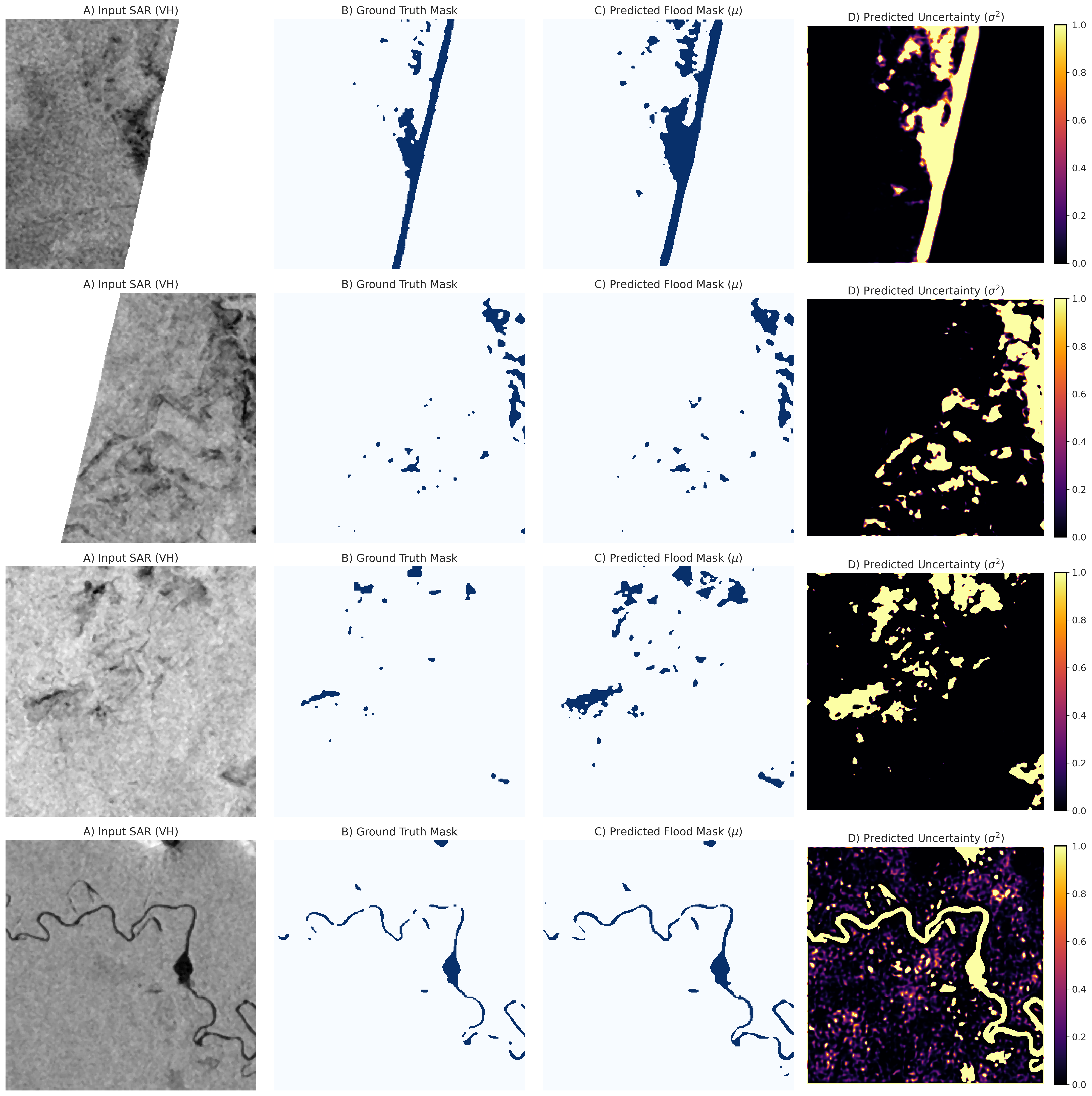}
    \caption{Flood prediction results visualization using a probabilistic framework. Each row corresponds to a different sample, with the following panels (from left to right): 
    \textbf{(A)} Input SAR (VH) data, \textbf{(B)} Ground truth flood mask, \textbf{(C)} Predicted flood mask ($\mu$), and \textbf{(D)} Predicted uncertainty map ($\sigma^2$). 
    Panel \textbf{(D)} highlights localized uncertainty, which is highest in regions of complex terrain or ambiguous radar signals, including radar shadows, fragmented water boundaries, and areas affected by speckle noise. These patterns of uncertainty reflect the heteroscedastic nature of the model, which adjusts its confidence based on the clarity of sensor data and physical consistency with the Shallow Water Equations.
    }
    \label{fig:uncertainty}
\end{figure*}

The superior quantitative performance of the probabilistic framework is, in fact, directly tied to its spatial interpretability, as visualized in Figure \ref{fig:uncertainty}. The multi-panel spatial maps reveal the internal mechanism by which the network negotiates the conflict between noisy sensor data and rigid physical laws. Crucially, the predicted aleatoric variance map ($\sigma^2$) demonstrates that the network does not distribute uncertainty uniformly. Instead, uncertainty is highly localized and context-aware.

As highlighted in the inset panels of Figure \ref{fig:uncertainty}, $\sigma^2$ peaks precisely in regions of known SAR ambiguity: along complex, fragmented water boundaries, within radar shadows cast by steep terrain or urban structures, and over open water surfaces corrupted by severe speckle due to wind roughening. This spatial localization is the physical manifestation of the heteroscedastic Negative Log-Likelihood objective. By predicting high variance in these specific pixels, the network mathematically down-weights their contribution to the total loss. Consequently, this proves that the model dynamically `relaxes' the strict enforcement of the Shallow Water Equations where the sensor data is deemed untrustworthy. Rather than forcing a physical fit onto optical illusions, the network learns to trust the physics primarily in clean, unambiguous regions, allowing the probabilistic formulation to seamlessly bridge the gap between noisy observations and hydrodynamic reality.

\subsection{Statistical Calibration of Uncertainty Estimates}

\begin{figure}[htbp]
    \centering
    \includegraphics[width=\columnwidth]{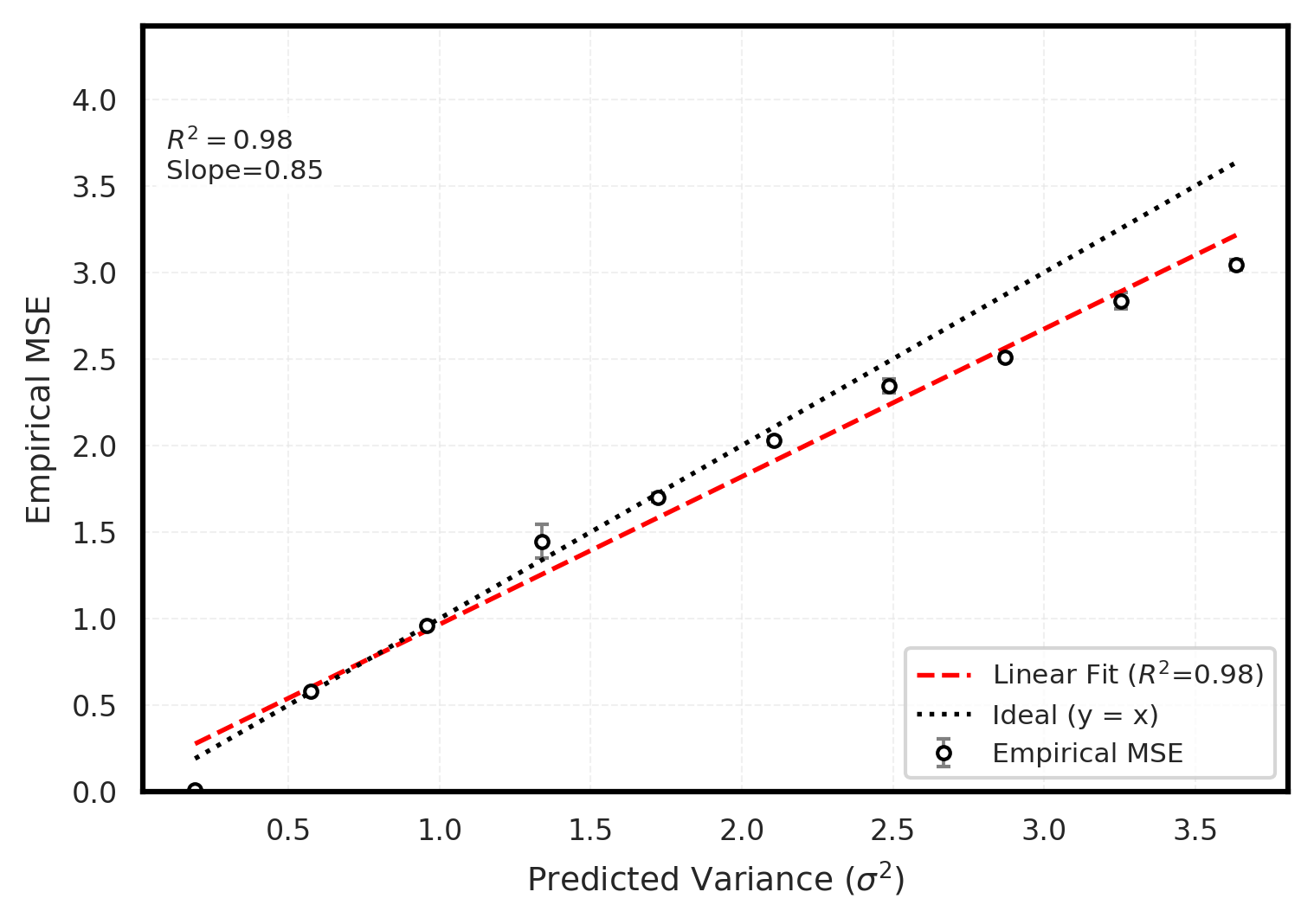}
    \caption{\textbf{Uncertainty Calibration and Error Correlation.} Empirical Mean Squared Error (MSE) of the predicted flood depth plotted against bins of the model's predicted aleatoric variance ($\sigma^2$). Error bars denote the standard error of the mean (SEM) within each bin. The strong positive correlation (Linear Fit $R^2 = 0.9794$) demonstrates that the probabilistic head is well-calibrated; regions where the network predicts high variance reliably correspond to areas of higher empirical predictive error. The dashed red line shows the linear fit, while the dotted line indicates ideal calibration ($y = x$).}
    \label{fig:uncertainty_calibration}
\end{figure}

While qualitative spatial mapping provides intuitive validation, operational deployment requires rigorous mathematical verification that the predicted uncertainty accurately reflects the model's true error distribution. In essence, this acts as a `self-awareness' test for the network: when the model claims to be uncertain, is it actually making larger errors? To quantify this, we conducted a statistical calibration analysis, as presented in Figure \ref{fig:uncertainty_calibration} and summarized in Table \ref{tab:calibration_stats}. We grouped the pixel-wise predictions across the validation set into bins based on their predicted aleatoric variance ($\sigma^2$). For each of these bins (represented as individual data points in the plot), we calculated the actual empirical Mean Squared Error (MSE) of the depth predictions.

The resulting linear regression reveals a remarkably strong, statistically significant positive correlation ($R^2 = 0.9794$, $p < 0.001$, slope $= 0.8534$). Across the predicted variance range, the empirical error closely tracks the predicted variance, approaching a one-to-one relationship. This near-unity slope is the precise signature of a well-calibrated model. It confirms that in regions where the network expresses high confidence (low variance), its depth predictions are indeed highly accurate; conversely, when it flags a region as highly uncertain, its empirical errors are verifiably larger.

This tight coupling, approaching the ideal $y = x$ relationship, demonstrates that the model is not arbitrarily generating variance values to escape the physics loss, nor is its uncertainty output merely random noise. Rather, it is producing highly calibrated, statistically sound estimates of its own unreliability. For disaster response agencies, this calibration is critical. It ensures that the network's probabilistic outputs can be directly translated into actionable risk assessments. For example, if the model flags a flooded urban sector with high aleatoric uncertainty due to severe radar double-bounce, decision-makers know not to blindly trust the depth estimates there, and can instead allocate secondary assets (such as drone reconnaissance) to verify the situation based on the verified certainty of the flood maps.

\begin{table}[htbp]
    \centering
    \caption{\textbf{Statistical Calibration Metrics.} Summary of the linear regression analysis evaluating the correlation between predicted aleatoric variance ($\sigma^2$) and empirical Mean Squared Error (MSE) of predicted flood depth.}
    \label{tab:calibration_stats}
    \begin{tabular}{l l}
        \toprule
        \textbf{Metric} & \textbf{Value} \\
        \midrule
        Coefficient of Determination ($R^2$) & 0.9794 \\
        Regression Slope & 0.8534 \\
        P-value & $4.96 \times 10^{-8}$ \\
        \bottomrule
    \end{tabular}
\end{table}

\subsection{Disentangling Data Noise from Model Ignorance}

\begin{figure*}[htbp]
    \centering
    \includegraphics[width=0.95\textwidth]{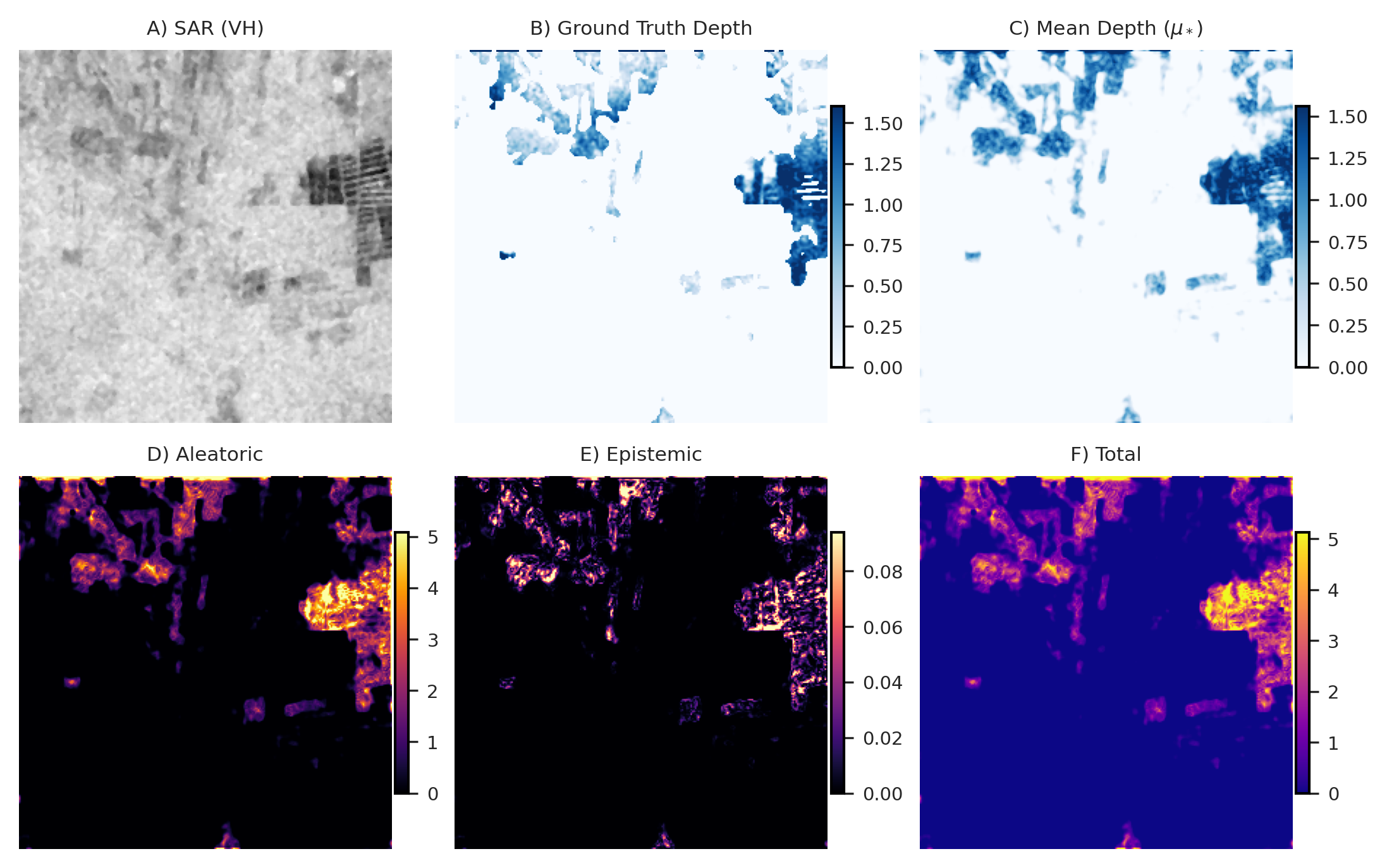}
    \caption{\textbf{Disentangling Aleatoric and Epistemic Uncertainty via Deep Ensembles.} The top row establishes the predictive context: \textbf{(A)} the noisy Input SAR (VH) signal, \textbf{(B)} the hydraulically simulated ground truth depth, and \textbf{(C)} the ensemble's mean predicted depth ($\mu_*$). The bottom row isolates the sources of doubt: \textbf{(D)} Aleatoric uncertainty successfully maps intrinsic sensor noise (e.g., speckle and radar shadow boundaries). \textbf{(E)} Epistemic uncertainty is minimal, indicating near-consensus across ensemble members. \textbf{(F)} The Total Predictive Variance, which is overwhelmingly dominated by the aleatoric data noise, confirming the model's structural robustness.}
    \label{fig:uncertainty_disentanglement}
\end{figure*}

Although the heteroscedastic probabilistic head successfully captures the intrinsic sensor noise of SAR, a complete uncertainty profile must also account for the network's own structural ignorance when faced with unfamiliar inputs. Figure \ref{fig:uncertainty_disentanglement} visualizes the successful disentanglement of these two distinct uncertainty sources using our Deep Ensemble methodology, governed by the Law of Total Variance. By decomposing the total predictive variance, we expose the underlying drivers of model ambiguity.

As summarized in Table \ref{tab:uncertainty_stats} and visualized in the second row of Figure \ref{fig:uncertainty_disentanglement}, the empirical results reveal a striking dynamic: the total predictive variance is overwhelmingly dominated by aleatoric uncertainty. The aleatoric component (Figure \ref{fig:uncertainty_disentanglement}D), calculated as the mean of the predicted variances across the ensemble members, isolates the irreducible noise, illuminating speckle-heavy regions, radar shadows, and complex water boundaries. 

Conversely, the epistemic uncertainty (Figure \ref{fig:uncertainty_disentanglement}E), calculated as the variance of the predicted means across the ensemble, remains very low ($1.18\%$ relative contribution). This minimal epistemic uncertainty is a strong indicator of model robustness; it demonstrates that the independent models within the ensemble have largely converged and exhibit near-consensus regarding the physical interpretation of the terrain. 

This explicit disentanglement provides immense operational value by eliminating diagnostic ambiguity. Because the epistemic uncertainty is negligible, researchers and practitioners can infer that collecting substantially more training data for this specific topography is unlikely to yield significant performance gains. The model is already highly confident in its learned physical rules. Instead, the high aleatoric uncertainty explicitly dictates that the SAR data itself is locally unreliable due to physical sensor limitations. In these specific flagged regions, confidence can only be improved through multi-modal sensor fusion (e.g., integrating optical data) or secondary reconnaissance, offering a highly actionable, physically grounded pathway for disaster response.

\begin{table}[htbp]
    \centering
    \caption{\textbf{Uncertainty Decomposition Statistics.} Quantitative breakdown of the total predictive variance into its aleatoric (data noise) and epistemic (model ignorance) components over the analyzed validation sample.}
    \label{tab:uncertainty_stats}
    \begin{tabular}{l c c}
        \toprule
        \textbf{Uncertainty Component} & \textbf{Mean} & \textbf{Standard Deviation} \\
        \midrule
        Aleatoric Variance ($\frac{1}{M}\sum \sigma^2_m$) & 0.3335 & 0.8887 \\
        Epistemic Variance ($\mathrm{Var}(\mu_m)$) & 0.0040 & 0.0140 \\
        Total Predictive Variance & 0.3377 & 0.8986 \\
        \midrule
        \textbf{Epistemic Contribution Ratio} & \multicolumn{2}{c}{\textbf{1.18\%}} \\
        \bottomrule
    \end{tabular}
\end{table}

\section{Discussion}

\subsection{Paradigm Shift in Earth Observation PINNs}
Prior attempts to stabilize Physics-Informed Neural Networks (PINNs) in multi-objective optimization have predominantly relied on math-centric dynamic weighting algorithms. While these techniques successfully balance competing loss gradients in pristine, simulated environments, they systematically fail when applied to raw satellite imagery because they treat all pixels as equally valid observations. Our framework introduces a fundamental paradigm shift toward sensor-centric stabilization. We demonstrate that for Earth Observation applications, sensor physics (the aleatoric noise inherent to the instrument) must dictate the application of mathematical physics (the governing hydrodynamic equations). By empowering the network to explicitly predict its own observational variance through a heteroscedastic loss function, our model localizes the relaxation of rigid physical constraints. This dynamic yielding prevents the `Physics Shock' that has historically bottlenecked PINN deployment in noisy remote sensing tasks. Furthermore, our comprehensive ablation study conclusively proves that `Smarter Beats Deeper.' In operational domains dominated by severe signal-to-noise limitations, such as SAR-based flood mapping, explicitly modeling uncertainty yields significantly higher performance and stability returns, demonstrated by a +25\% relative IoU improvement, than blindly increasing the architectural depth or representational capacity of deterministic models, as seen with the deeper Res-UNet.

\subsection{Operational Implications for Disaster Response}
In the context of disaster response and humanitarian relief, deterministic `flood/no-flood' binary maps are frequently insufficient, and occasionally dangerous, for decision-making. Standard deterministic models fail silently; they provide no indication of whether a misclassification stems from a minor algorithmic error or a complete sensor failure. Our uncertainty-aware framework resolves this critical operational vulnerability by providing calibrated confidence intervals alongside every prediction. By successfully disentangling uncertainty, we equip emergency response agencies with a nuanced, actionable risk assessment tool. Decision-makers can now quantitatively distinguish between regions where the sensor is fundamentally blind, such as radar shadows in steep valleys or severe speckle over wind-roughened water (aleatoric uncertainty), and regions where the model simply lacks adequate training context (epistemic uncertainty). This calibrated confidence ensures that rescue resources are not misallocated based on radiometric artifacts, fundamentally upgrading the utility of deep learning from a perceptual mapping tool to a reliable operational inference engine.

\subsection{Limitations, Boundaries and Future Work}

While highly effective at producing robust and physically interpretable flood maps, this framework introduces specific operational and data dependencies that must be acknowledged. First, the rigorous disentanglement of aleatoric and epistemic uncertainty relies fundamentally on Deep Ensembles. Training and running inference on multiple independent models (e.g., $M=3$ or $M=5$) linearly increases the computational overhead and memory footprint. This associated inference latency may pose significant challenges for ultra-low-latency or resource-constrained edge deployment scenarios where rapid, single-pass inference is prioritized. Second, the physical validity of the model is inextricably linked to the quality of the auxiliary topographic data. The framework explicitly relies on a high-fidelity Digital Elevation Model (DEM) to compute the spatial derivatives required for the Water Surface Elevation (WSE) smoothness constraint. Consequently, the framework's physical reasoning is bounded by the vertical accuracy and spatial resolution of the provided DEM. Any inherent artifacts, severe vertical errors, or temporal misalignments in the DEM will inevitably propagate into the physics loss, potentially misleading the optimization process and degrading the final inundation estimate.

To address the computational constraints identified above, future research should explore computationally lighter alternatives to Deep Ensembles for estimating epistemic uncertainty. Techniques such as Evidential Deep Learning, Monte Carlo Dropout, or Laplace approximations offer promising avenues to maintain uncertainty awareness while significantly reducing inference latency. Furthermore, the concept of uncertainty gating can be naturally extended to multi-modal sensor fusion. Future architectures could employ the predicted aleatoric variance as a dynamic, pixel-wise gating mechanism to automatically shift reliance between SAR and optical imagery (e.g., Sentinel-2), depending on local noise characteristics like cloud cover or radar shadow. Most importantly, having established a numerically stable PINN framework that does not collapse under Earth Observation noise, the foundation is now laid for a more profound application of these models. 

\section{Conclusion}

The integration of physical laws into Deep Learning represents the next frontier in Earth Observation, promising models that are both scalable and scientifically consistent. However, the inherent noise of satellite sensors like SAR fundamentally conflicts with rigid mathematical constraints, leading to optimization failures. In this study, we demonstrated that the solution to this `Physics Shock' does not lie in simply increasing network depth or capacity, but in rigorous probabilistic modeling.

By treating the flood mapping task as an uncertainty-aware optimization problem, our framework successfully bridged the gap between noisy sensor data and the Shallow Water Equations. The introduction of heteroscedastic aleatoric uncertainty allowed the model to act as a dynamic gate, enforcing physics where the data is clear and relying on probabilistic variance where the sensor is blinded by speckle or shadow. Coupled with a Warm-Start conditioning protocol, this approach entirely stabilized the training gradients, yielding a physically consistent model that outperformed deterministic baselines by +25\% in IoU.

Ultimately, this work proves that for Physics-Informed Neural Networks to succeed in remote sensing, sensor physics must guide mathematical physics. Providing calibrated confidence intervals by disentangling aleatoric and epistemic uncertainty ensures that these models can be safely deployed for critical disaster response. Having established a numerically stable foundation, future research can now pivot from enforcing known physical constants to leveraging these networks for inverse physics discovery in uncalibrated environments.

\section*{Acknowledgment}

\noindent This research article has been made possible partly with the support of National Aeronautics and Space Administration (NASA) award number 80NSSC23M0051 and in part by the National Science Foundation (NSF) Grant under Award 2401942.

\bibliographystyle{IEEEtran}
\bibliography{refs}

\begin{IEEEbiography}[{\includegraphics[width=1in,height=1.25in,clip,keepaspectratio]{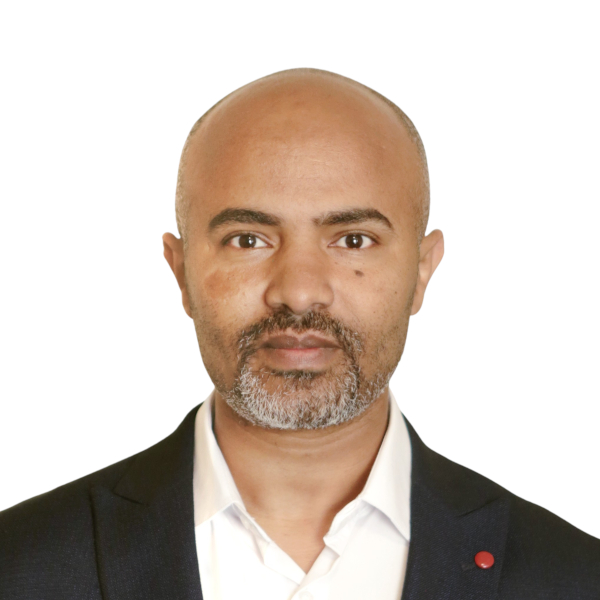}}]{Tewodros Syum Gebre}
holds a B.S. in Hydraulic Engineering from Arba Minch University, Ethiopia, in 2009, an M.S. in Civil (Road and Transport) Engineering from Addis Ababa University, Ethiopia, in 2014, and a Ph.D. in Applied Science and Technology at North Carolina A\&T State University, USA, in 2025. His expertise spans developing machine learning models for automating traffic management, image and both structure and unstructured data analysis, and system development. 

He is currently a Postdoctoral Researcher at North Carolina A\&T State University, in AI and ML AI applications for smart cities and Infrastructure Resilience. He worked on a Microsoft-funded project developing AI-based traffic monitoring systems using transformer models and UAV imagery. He is a former member of the NC-CAV Center, supported by the North Carolina Department of Transportation. He received the G. Herbert Stout Award for Best Student Paper at the North Carolina GIS (NCGIS) Conference and was awarded the Excellence Scholarship from 2021 to 2024. He also received Microsoft’s AFMR grant for the 2023–2024 period.
\end{IEEEbiography}

\begin{IEEEbiography}[{\includegraphics[width=1in,height=1.25in,clip,keepaspectratio]{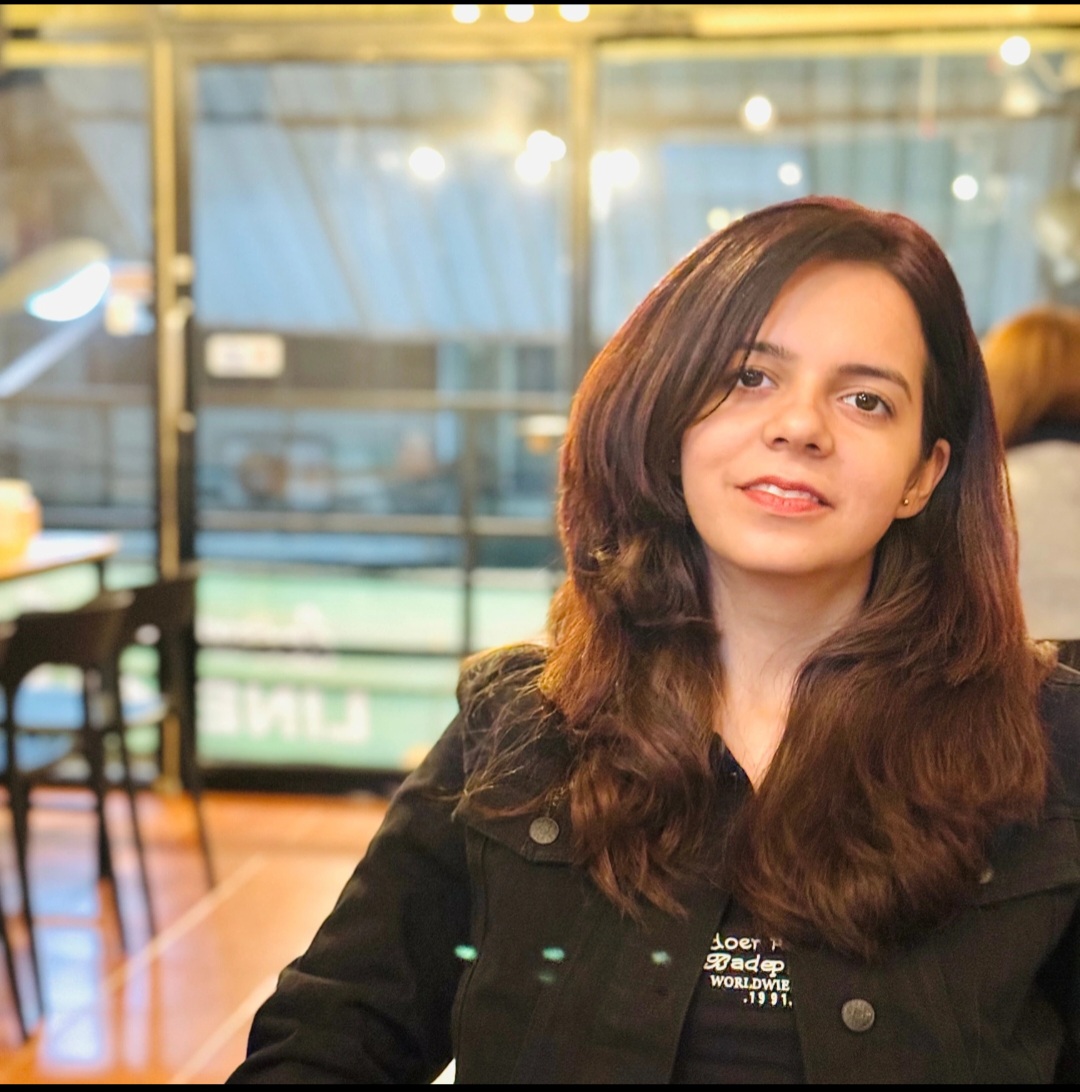}}]{Jagrati Talreja} (Graduate Member, IEEE) received the B.Tech. degree in Electronics and Communication Engineering (Networks) from Pranveer Singh Institute of Technology, Kanpur, Uttar Pradesh, India, in 2019. She pursued and successfully completed a five-year integrated Ph.D. in Electrical Engineering at Chulalongkorn University, Bangkok, Thailand, in 2024. Dr. Jagrati is currently a Postdoctoral Researcher at North Carolina A\&T State University, Greensboro, NC, USA. She is working on projects funded by NASA and the National Science Foundation (NSF), focusing on the development and application of deep learning algorithms and advanced geospatial techniques for analyzing satellite and UAV imagery in flood mapping and disaster assessment.

Her research interests include Electrical Engineering, Remote Sensing, Satellite Image Processing, Neural Networks, and Machine Learning, with a particular emphasis on deep learning for image super-resolution, multimodal data fusion, and SAR-to-optical image translation. She has hands-on expertise in designing hybrid CNN–Transformer architectures, GAN-based frameworks, and transfer learning models for geospatial applications.
\end{IEEEbiography}

\begin{IEEEbiography}[{\includegraphics[width=1in,height=1.25in,clip,keepaspectratio]{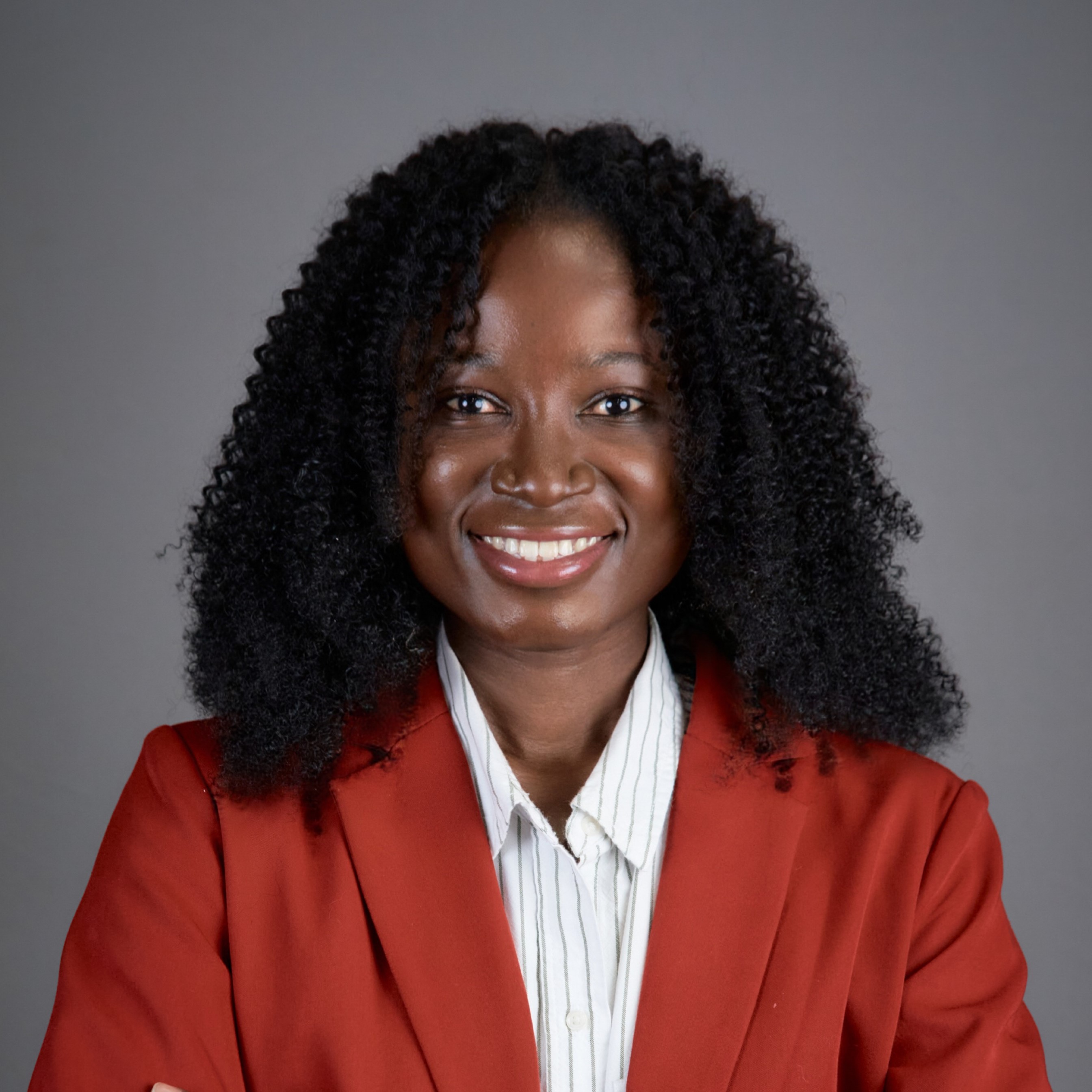}}]{Matilda Anokye} received the B.S. degree in Geomatic Engineering from the University of Mines and Technology, Ghana, in 2018, and the M.S. degree in Urban Forestry and Natural Resources from Southern University A\&M College, Baton Rouge, LA, in 2022. She is currently pursuing the Ph.D. degree in Applied Science and Technology at North Carolina A\&T State University, Greensboro, NC, working on a NASA and NOAA-funded project that utilizes UAV, LiDAR, SAR, and optical imagery to model flood-induced inundation patterns and assess wetlands and watersheds.

Her research interests and skills span the area of data fusion, GeoAI, hydrological modeling, LiDAR point cloud processing, spatial analytics, and the application of data science and AI approaches to analyze remote sensing data for environmental monitoring and geospatial data-driven analysis.
\end{IEEEbiography}

\begin{IEEEbiography}[{\includegraphics[width=1in,height=1.25in,clip,keepaspectratio]{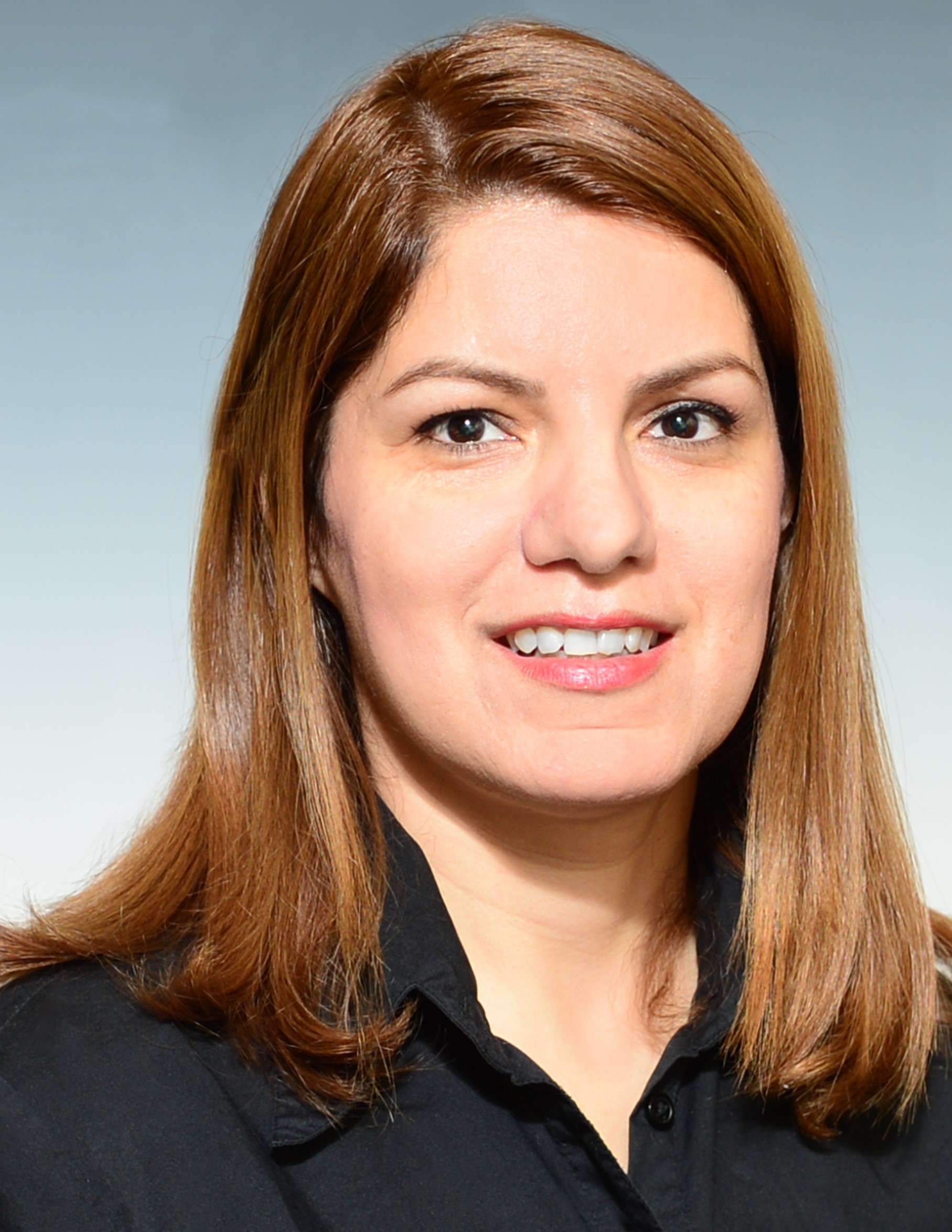}}]{Leila Hashemi-Beni} is an associate professor and the director of the NASA-funded institute for Harnessing Data Science for Environmental Management at North Carolina A\&T State University. She holds a B.S. in Civil-Surveying Engineering (Geomatics) from the University of Isfahan, followed by an M.S. in Civil-Surveying Engineering (Photogrammetry/Remote Sensing) from the University of Tehran, and a Ph.D. in Geospatial Information System from Laval University.

Her research experience and interests span the areas of geospatial data science, UAV and satellite remote sensing, multi-temporal and multisource data fusion and image classification, 3D data modeling, automatic matching and change detection between various datasets, and developing GIS and remote sensing methodologies for environmental management. She is currently working as a PI/Co-PI on many projects supported by NASA, NSF, NOAA, Microsoft, North Carolina Collaboratory, and North Carolina DoT. She has served as a proposal panelist and reviewer for many US and international funding organizations, and she has served as a chair/co-chair or as a scientific committee member for many national or international conferences/workshops. She is currently serving as the Co-Chair of the LiDAR, Laser Altimetry, and Sensor Integration Working Group, International Society of Photogrammetry and Remote Sensing (ISPRS). She is the PI of the project `AI-based traffic monitoring systems using generative pre-trained transformer models and high-resolution UAV imagery' supported by Microsoft’s Accelerating Foundation Models Research program.
\end{IEEEbiography}

\end{document}